%% file: iclr2022_conference.tex
\documentclass{article} 
\usepackage{iclr2022_conference,times}

\input{math_commands.tex}

\usepackage{hyperref}
\usepackage{url}
\usepackage[english]{babel}
\usepackage{graphicx}
\usepackage{wrapfig}
\usepackage{float}
\usepackage{algorithm}
\usepackage{algorithmic}
\usepackage{booktabs}
\usepackage{subfigure}

\title{Transfering Hierarchical Structure with Dual Meta Imitation Learning}

\author{Chongkai Gao, Yizhou Jiang, Feng Chen
\\Department of Automation\\
Tsinghua University\\
Beijing, China 100086 \\
\texttt{\{gck20,jyz20\}@mails.tsinghua.edu.cn} \texttt{\{chenfeng\}@mail.tsinghua.edu.cn}\\
}

\iclrfinalcopy 
\begin{document}

\maketitle

\begin{abstract}

Hierarchical Imitation Learning (HIL) is an effective way for robots to learn sub-skills from long-horizon unsegmented demonstrations. However, the learned hierarchical structure lacks the mechanism to transfer across multi-tasks or to new tasks, which makes them have to learn from scratch when facing a new situation. Transferring and reorganizing modular sub-skills require fast adaptation ability of the whole hierarchical structure. In this work, we propose Dual Meta Imitation Learning (DMIL), a hierarchical meta imitation learning method where the high-level network and sub-skills are iteratively meta-learned with model-agnostic meta-learning. DMIL uses the likelihood of state-action pairs from each sub-skill as the supervision for the high-level network adaptation, and use the adapted high-level network to determine different data set for each sub-skill adaptation. We theoretically prove the convergence of the iterative training process of DMIL and establish the connection between DMIL and Expectation-Maximization algorithm. Empirically, we achieve state-of-the-art few-shot imitation learning performance on the Meta-world \cite{metaworld} benchmark and competitive results on long-horizon tasks of Kitchen environments.

\end{abstract}

\section{Introduction}

Imitation learning (IL) has shown promising results for intelligent robots to conveniently acquire skills from expert demonstrations \cite{reinforceandimitation,deepmimic}. Nevertheless, imitating long-horizon unsegmented demonstrations has been a challenge for IL algorithms, because of the well-known issue of compounding errors \cite{areductionofimitation}. This is one of the crucial problems for the application of IL methods to robots since plenty of practical manipulation tasks are long-horizon. Hierarchical Imitation Learning (HIL) aims to tackle this problem by decomposing long-horizon tasks with a hierarchical model, in which a set of sub-skills are employed to accomplish specific parts of the long-horizon task, and a high-level network is responsible for determining the switching of sub-skills along with the task. Such a hierarchical structure is usually modeled with Options \cite{ddo,ddco,optiongail} or goal-conditioned IL paradigms \cite{hierarchicalimitationlearning}. HIL expresses the nature of how humans solve complex tasks, and has been considered to be a valuable direction for IL algorithms \cite{analtorithmicperspectiveonimitationlearning}.

\begin{figure*}[t]
	\centering  
	\subfigure[Illustration of the bi-level transfer problem of HIL in new tasks.]{
		\label{intro}
		\includegraphics[width=0.585\textwidth]{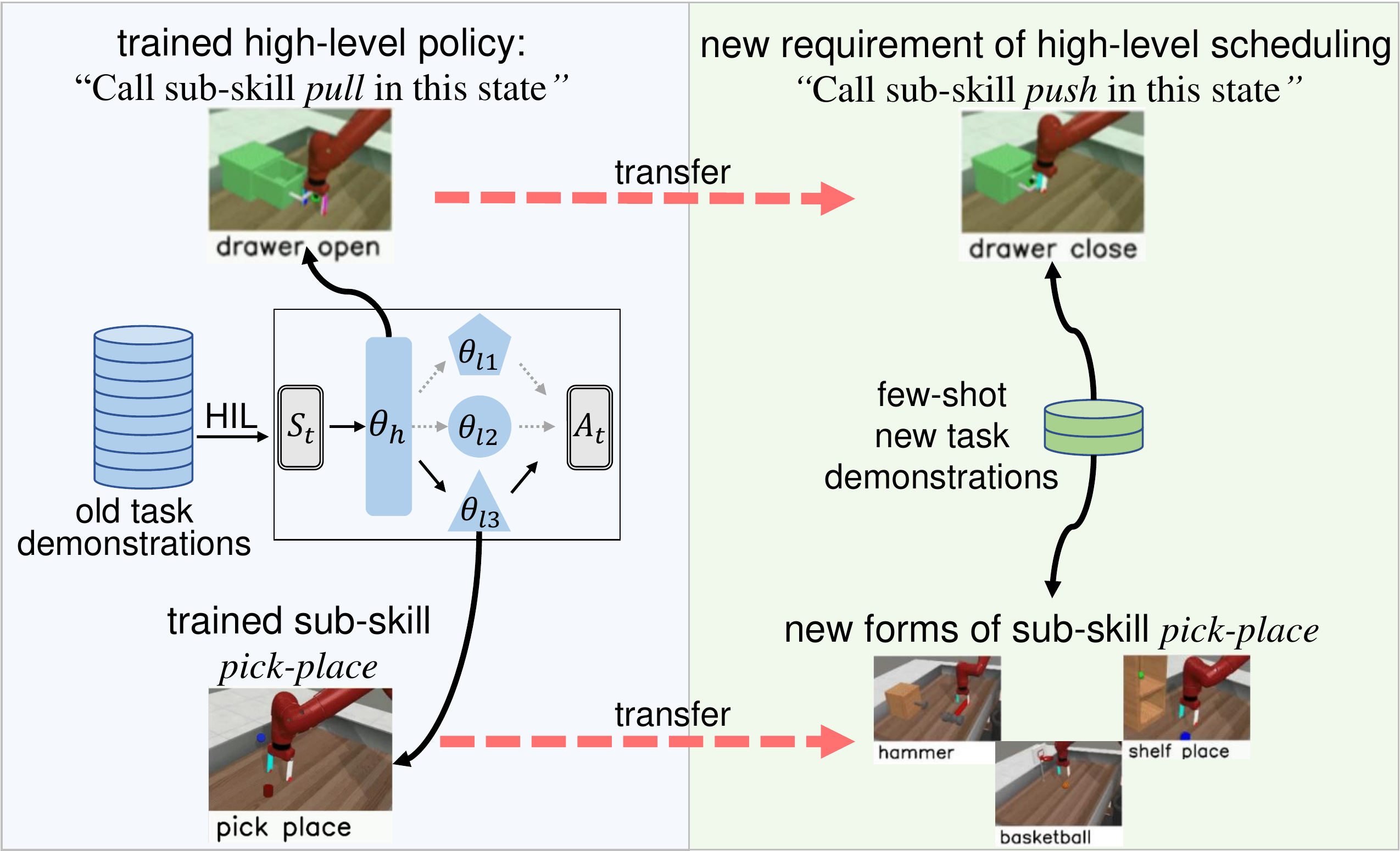}}
	\subfigure[Comparison of MIL and DMIL.]{
		\label{intro2}
		\includegraphics[width=0.385\textwidth]{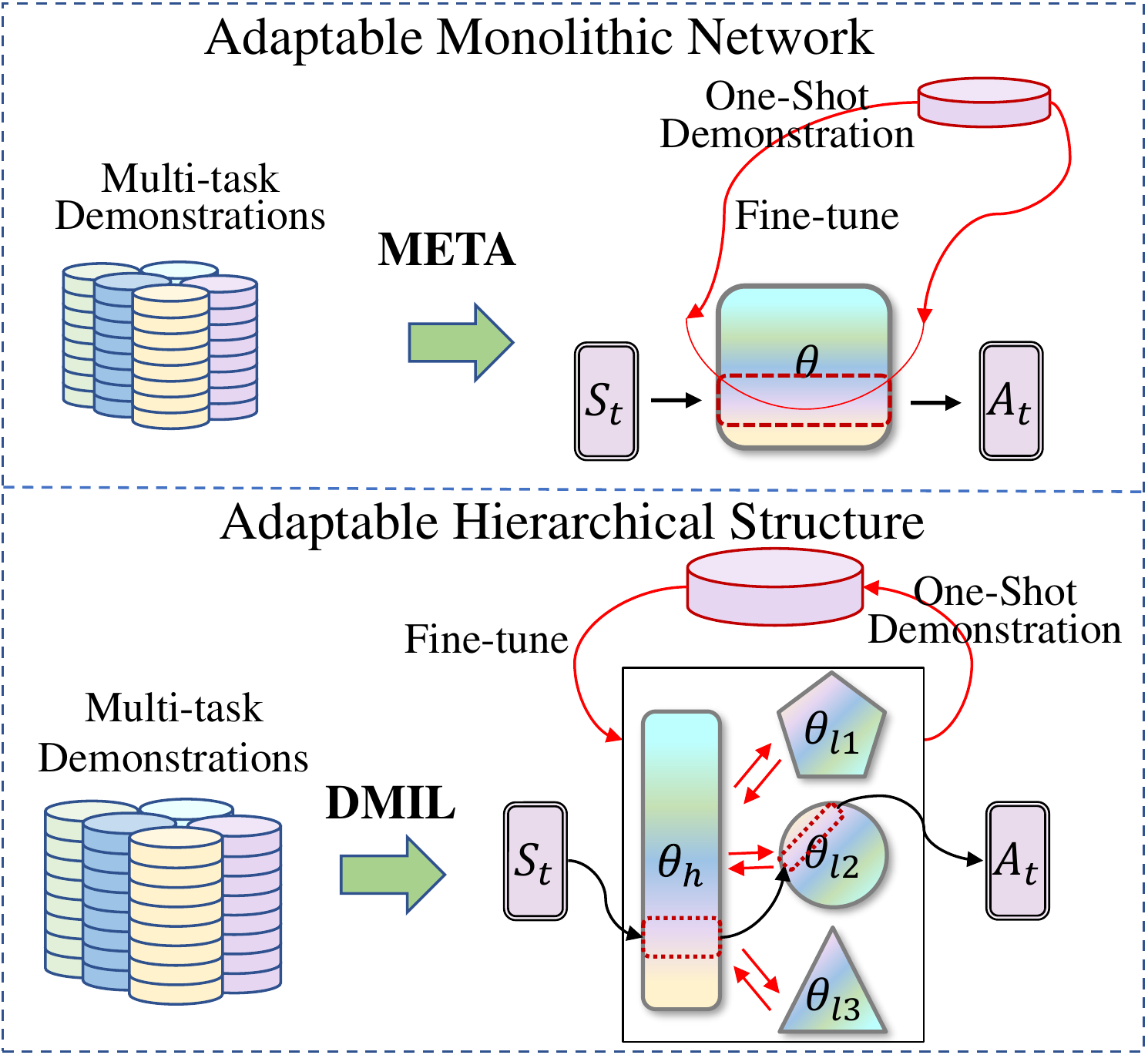}}
	
	\vspace{-0.12in}
	
	\caption{(a) Both the high-level network and sub-skills need to be transferred to new tasks. Above: when the robot arm is over a half-open drawer, the task can be either opening or closing the drawer, which requires the high-level network to call different sub-skills. Below: a same sub-skill \textit{pick-place} may exhibit different specific forms in new tasks. (b) DMIL aims to integrate MAML into HIL with a novel iterative optimization procedure that meta-learns both the high-level network and sub-skills.}
	\label{Fig.main}
	\vspace{-0.12in}
\end{figure*}

However, most current HIL methods have no explicit mechanism to transfer previously learned sub-skills to new tasks with few-shot demonstrations. This requirement comes from that the learned hierarchical structure may conflict with discrepant situations in new tasks. As shown in \ref{intro}, both the high-level network and sub-skills need to be transferred to new forms to satisfy new requirements: the high-level network needs new manners to schedule sub-skills in new tasks (for example, calling different sub-skills at the same state), and each sub-skill needs to adapt to new specific forms in new tasks (for example, grasping different kinds of objects). An appropriate approach needs to be exploited to endow HIL with such kind of ability to simultaneously transfer both the high-level network and sub-skills with few-shot new task demonstrations, and this can significantly increase the generalization ability of HIL methods and make them be applied to a wider range of scenarios.

Recently, meta imitation learning (MIL) \cite{oneshotimitationfromobservinghumans,oneshotimitationvisuallearningviametalearning,oneshothierarchicalimitationlearning} employ model-agnostic meta-learning (MAML) \cite{maml} into imitation learning procedure to enable the learned policy to quickly adapt to new tasks with few shot demonstrations. MAML first fine-tunes the policy network in the inner loop, then evaluates the fine-tuned network to update its initial parameters with end-to-end gradient descent at the outer loop. The success of MIL inspires us to integrate MAML into HIL to transfer the hierarchical structure in new tasks. However, this is not straightforward. The reason why MAML can be directly applied to IL methods is that most IL methods learn a monolithic policy in an end-to-end fashion, which conforms to the original setting of MAML. However, in HIL, the bi-level policy network is trained in an iterative and self-supervised paradigm. Intuitively, this bi-level optimization procedure makes it necessary to add MAML into both levels of HIL to transfer the whole structure. Since MAML itself is also a bi-level process, this becomes a difficult problem to schedule the complex inner loops and outer loops of meta-learning processes of the bi-level networks in HIL to ensure the training convergence.


In this work, we propose a novel hierarchical meta imitation learning framework called Dual Meta Imitation Learning (DMIL) to successfully incorporate MAML into the iterative training process of HIL, as shown in \ref{intro2}. We firstly adopt the EM-like HIL method \cite{ddo} as our basic bi-level HIL structure, where the high-level network and sub-skills are in mutual supervision: the likelihood of each state-action pair in few-shot demonstrations of each sub-skill can provide supervisions for the training of high-level network, and the high-level network in turn determines the data sets for the training of each sub-skill. Then we design an elaborate bi-level MAML procedure for this hierarchical structure to make it can be fully meta-learned. In this procedure, we first fine-tune the high-level network and sub-skills \textit{in sequence} at inner loops, then meta-update them \textit{simultaneously} at outer loops. We theoretically prove the convergence of this special training procedure by leveraging previous results from \cite{amortizedbayesianmeta,pmaml,gradientem} to reframe both MAML and DMIL as  hierarchical Bayes inference processes and get the convergence of DMIL according to the convergence of MAML from previous results \cite{convergenceofmaml}.

We test our method on the challenging meta-world benchmark environments \cite{metaworld} and the Kitchen environment of D4RL benchmarks \cite{d4rl}. In our experiments, we successfully acquire a set of meaningful sub-skills from a large scale of manipulation tasks, and achieve state-of-the-art few-shot imitation learning abilities in the ML45 suite. In summary, the main contributions of this paper are as follows:
\begin{itemize}
	\item We propose DMIL, a novel hierarchical meta imitation learning framework that meta-learns both the high-level network and sub-skills from unsegmented multi-task demonstrations in a general EM-like fashion.
	
	\item We propose a novel training algorithm for DMIL to schedule the meta-learning processes of its bi-level networks and theoretically prove its convergence.
	
	\item We achieve state-of-the-art few-shot imitation learning performance on meta-world benchmark environments and competitive results in the Kitchen environment.
\end{itemize}

\section{Related Work}

\subsection{Hierarchical Imitation Learning}

Recovering inherent sub-skills  contained in expert demonstrations and then reuse them with hierarchical structures has long been an interesting topic in the hierarchical imitation learning (HIL) domain. According to whether there are pretraining tasks, we can divide HIL methods into two categories. The first kind  aims to manually design a set of simple pretraining tasks that could encourage distinct skills or primitives, and then they learn a high-level network to master the switching of primitives to accomplish complex tasks \cite{stochasticnn,mcp,pretrain1,pretrain2,hierarchicalimitationlearning}. However, for unsegmented  demonstrations where no pretraining tasks are provided, which is the situation in our paper, these methods can not be applied.

The second kind of methods  aim to learn sub-skills with unsupervised  learning methods. \citet{ddo,ddco} acquire Options \cite{option} from  demonstrations with an Expectation-Maximization-like procedure and use the Baum-Welch algorithm to estimate the parameters of different options. \citet{optiongan,optiongail} integrate  generative adversarial networks into option discovery process. \citet{infogail,directedinfogail,learningcompoundtaskswithouttaskspecificknowledge} incorporate generative-adversarial imitation learning \cite{gail} framework and an information-theoretic metric \cite{infogan} to simultaneously imitate the expert and maximize the mutual-information between latent sub-skill categories and corresponding trajectories to acquire decoupled sub-skills. There are some methods called mixture-of-expert (MoE) that compute the weighted sum of all primitives to get the action rather than only using one of them at each time step  \cite{moe1,moe2,moe3}. Other methods aim to seek an appropriate latent space that can map sub-skills into it and then condition a policy on the latent variable to reuse sub-skills  \cite{latentplay,latentspace,learninganembeddingspace,fist,spirl}. 

For transferring learned sub-skills, some work fine-tune the whole structure in new tasks \cite{fist}. However, the performance of fine-tuning all depends on the generalization of deep networks, which may vary among different tasks and network designs. 


\subsection{Meta Imitation Learning}

Meta imitation learning, or one-shot imitation learning, leverages various meta-learning methods and multi-task demonstrations to meta-learn a policy that can be quickly adapted to a new task with few-shot new task demonstrations. \citet{oneshotimitationlearning,transformerbasedmeta} employ self-attention modules to process the whole demonstration and the current observation to predict current action. \citet{oneshothierarchicalimitationlearning,oneshotimitationvisuallearningviametalearning,oneshotimitationfromobservinghumans} use model-agnostic meta-learning (MAML) \cite{maml} to achieve one-shot imitation learning ability for various manipulation tasks with robot or human visual demonstrations.  \citet{learningaprioroverintent,metairl} propose to meta-learn a robust reward function that can be quickly adapted to new tasks and then use it to perform IRL in new tasks. However, they need downstream inverse reinforcement learning after the adaptation of reward functions, thus conflicts with our goal of few-shot adaptation. Most of above methods only learn one monolithic policy, lacking the ability to model multiple sub-skills in long-horizon tasks. Some works aim to tackle the multi-modal data problem in meta-learning by avoiding single parameters initialization across all tasks  \cite{multimodalmaml,modularmetalearning,metalearningsharedhierarchies,hierarchicallystructuredmetalearning}, but they lack the mechanism to schedule the switching of different sub-skills over time. There are some works that also meta-learn a set of sub-skills in a hierarchical structure \cite{oneshothierarchicalimitationlearning,metalearningsharedhierarchies}, but they either use manually designed pretraining tasks or relearn the high-level network in new tasks, which is not appropriate in few-shot imitation learning settings. 


\section{Method}

\label{method}

\subsection{Preliminaries}

We denote a discrete-time finite-horizon Markov decision process (MDP) as a tuple $ (\mathcal{S},\mathcal{A},T,P,r,\rho_0) $, where $ \mathcal{S} $ is the state space, $ \mathcal{A}  $ is the action space, $ T $ is the time horizon, $ P:\mathcal{S} \times \mathcal{A} \times \mathcal{S} \rightarrow [0, 1] $ is the transition probability distribution, $ r:\mathcal{S} \times \mathcal{A} \rightarrow \mathbb{R} $ is the reward function, and $ \rho_0 $ is the distribution of the initial state $ s_0 $. 

\begin{figure*}[t]
	\centering
	\includegraphics[width=\linewidth]{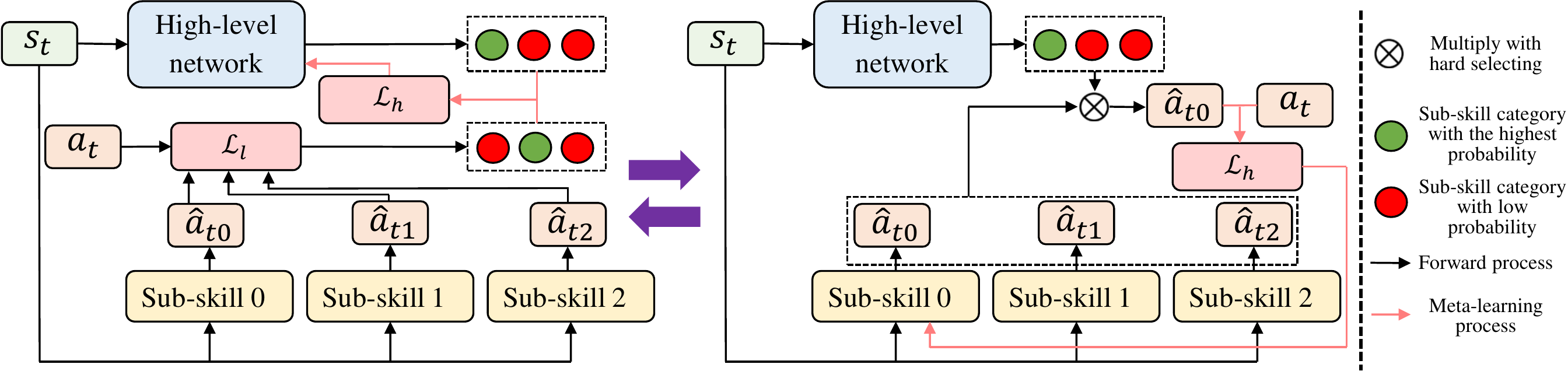}
	\vspace{-20pt}
	\caption{The iterative meta-learning process of DMIL at each iteration. Left: the supervision of high-level network (sub-skill categories) comes from the most accurate sub-skill (the green one, sub-skill 1 here). Right: the sub-skill updated at current step (the green one, sub-skill 0 here) is determined by the fine-tuned high-level network.\label{network}}
\end{figure*}

\subsection{Formulation of Meta Imitation Learning Problem}

We firstly introduce the general setting of the meta imitation learning problem. The goal of meta imitation learning is to extract some common knowledge from a set of robot manipulation tasks $ \{\mathcal{T}_i\} $ that come from the same task distribution $ p(\mathcal{T}) $, and adapt it to new tasks quickly with few shot new task demonstrations. As in model-agnostic meta-learning algorithm (MAML) \cite{maml}, we formalize the common knowledge as the initial parameter $ \theta $ of the policy network $ \pi_\theta $ that can be efficiently adapted
with new task gradients.  

For each task $ \mathcal{T}_i \sim p(\mathcal{T})$, a set of demonstrations $ \mathcal{D}_i $ is provided, where $ \mathcal{D}_i $ consists of $ N $ demonstration trajectories: $ \mathcal{D}_i=\left\{\tau_{ij}\right\}_{j=1}^{N} $, and $ \tau_{ij} $ consists of a sequence of state-action pairs: $ \tau_{ij}=\left\{(s_{t}, a_{t})\right\}_{t=1}^{T_{ij}} $, where $ T_{ij} $ is the length of $ \tau_{ij} $. Each $ \mathcal{D}_i $ is randomly split into support set $ \mathcal{D}_i^{tr} $ and query set $ \mathcal{D}_i^{val} $ for meta-training and meta-testing respectively. During the training phase, we sample $ m $ tasks from $ p(\mathcal{T}) $, and in each task $ \mathcal{T}_i $, we use $ \mathcal{D}_i^{tr} $ to fine-tune $ \pi_\theta $ to get the adapted task-specific parameters $ \lambda_i $ with gradient descent, and then evaluate it with $ \mathcal{D}_i^{val} $ to get the meta-gradient of $ \mathcal{T}_i $, and we optimize the initial parameters $ \theta $ with the average of meta-gradients from all $ m $ tasks. As in \citet{oneshotimitationvisuallearningviametalearning,oneshotimitationlearning}, we use behavior cloning \cite{bc} loss as our metrics for meta-training and meta-testing. It aims to train a policy $ \pi_\theta: \mathcal{S}\rightarrow\mathcal{A} $ that maximizes the likelihood such that $ \theta^*= \mathop{\arg\max}_{\theta}\sum_{i=1}^{N}\log \pi_\theta(a_i|s_i)$, where $ N $ is the number of provided state-action pairs. We denote the loss function of this optimization problem as $ \mathcal{L}_{BC}(\theta,\mathcal{D}) $, and the general objective of meta imitation learning problem is:
\begin{equation}
	\label{metaloss}
	\min \limits_{\theta} \sum_{i=1}^{m} \mathcal{L}_{BC}\left(\lambda_i, \mathcal{D}_i^{\text {val }}\right),
\end{equation}
where $ \lambda_i = \theta-\alpha\nabla_\theta\mathcal{L}_{BC}(\theta, \mathcal{D}_i^{tr}) $, and $ \alpha $ is a hyper-parameter which represents the inner-update learning rate.

\subsection{Dual Meta Imitation Learning (DMIL)}

In this work we assume at each time step $ t $, the robot may switch to different sub-skills to accomplish the task. We define the sub-skill category at each time step $ t $ as $ z_t=1,\cdots,K $, where $ K $ is the maximum number of sub-skills. We assume a successful trajectory $ \tau_{ij} $ of a task $ \mathcal{T}_i $ is generated from several (at least one) sub-skill policies, i.e., $ \tau_{ij} = \sum_{t=1}^{T_{ij}}\{(s_t, \pi_E(s_t|z_t))\} $, where $ \pi_E $ represents the expert policy. Our goal is to learn a hierarchical structure from multi-task demonstrations $ \{\mathcal{D}_1,\cdots,\mathcal{D}_m\} $ in an unsupervised fashion. In our model, a high-level network $ \pi_{\theta_h} $ that parameterized by $ \theta_h $ determines the sub-skill category $ \hat{z_t} $ at each time step $ t $, and the $ z $\textit{-th} sub-skill among $ K $ different sub-skills $ \pi_{\theta_{l1}},\cdots,\pi_{\theta_{lK}} $ will be called to predict the corresponding action $ \hat{a_t} $ of state $ s_t $, where the hat symbol denotes the predicted result. We use $ \lambda_h $ and $ \lambda_{l1},\cdots,\lambda_{lk} $ to represent the adapted parameters of $ \theta_h $ and $ \theta_{l1},\cdots,\theta_{lK} $ respectively. In order to achieve few-shot learning ability in new tasks, we condition the high-level network only on states, i.e., $ \hat{z_t} = \pi_{\theta_h}(s_t) $, since at the testing phase we only have access to states and have no access to action information. 

DMIL aims to first fine-tune both $ \pi_{\theta_h} $ and $ \pi_{\theta_{l0}},\cdots,\pi_{\theta_{lK}} $ and then meta-update them. In a new task, $ \pi_{\theta_h} $ may not provide correct sub-skill categories as stated in the introduction. However, sub-skills still retain the ability to give out supervision for the high-level network with knowledge learned from previously learned tasks and few-shot demonstrations. This is because most robot manipulation tasks are made up of a set of shared basis skills like \textit{reach, push} and \textit{pick-place}. As shown in the left side of \ref{network}, the sub-skill that gives out the closet $ \hat{a_t} $ to $ a_t $ can be seen as supervision for the high-level network to classify $ s_t $ into this sub-skill. On the other hand, the adapted high-level network can classify each data point in provided demonstrations to different sub-skills for them to perform fine-tuning, as shown in the right side of \ref{network}. In summary, DMIL contains four steps for one training iteration. We call them \textbf{High-Inner-Update (HI)}, \textbf{Low-Inner-Update (LI)}, \textbf{High-Outer-Update (HO)}, and \textbf{Low-Outer-Update (LO)}, which represents the fine-tuning and meta-updating process of the bi-level networks respectively. The key problem is how to arrange these optimization orders to ensure convergence. We first introduce these four steps formally here, then discuss how to schedule them in the next section. The whole procedure is summarized in algorithm \ref{alg}.

\textbf{HI:} For each sampled task $ \mathcal{T}_i $, we sample the first batch of trajectories $ \{\tau_{i1}\} $ from $ \mathcal{D}_i^{tr} $. The principle of this step is to use sub-skill that can predict the closest action to the expert action to provide self-supervised category ground truths for the training of high-level network, which is a classifier in form. We make \textit{every} state-action pair passed directly to each sub-skill and compute $ \mathcal{L}_{BC}(\theta_{lk},\tau_{i1}), k=1,\cdots,K $, and choose the ground truth at each time step as the sub-skill category $ k $ that minimizes $ \mathcal{L}_{BC}(\theta_{lk}, (s_t,a_t)) $ :
\begin{equation}
	\label{highlabel}
	p(z_{i1t}=k) =\left\{
	\begin{array}{lr}
		1, \ \text{if} \  k=\mathop{\arg\min}_{k}\mathcal{L}_{BC}(\theta_{lk}, (s_t,a_t)) &  \\
		0, \ \text{else} &  
	\end{array}
	\right.
	.
\end{equation}
Then we get predicted sub-skill categories from the high-level network: $ \hat{z_{i1t}}=\pi_{\theta_h}(s_t) $, and use a cross-entropy loss to train the high-level network:
\begin{equation}
	\label{losshighadapt}
	\mathcal{L}_h(\theta_h,\tau_{i1}) = -\frac{1}{T_{i1}}\sum_{t=1}^{T_{i1}}\sum_{k=1}^{K}p(z_{i1t}=k)\log p(\hat{z_{i1t}}=k).
\end{equation}
Finally we perform gradient descent on the high-level network and get $ \lambda_h = \theta_h-\alpha\nabla_{\theta_h}\mathcal{L}_h(\theta_h,\tau_{i1}) $. Note $ \theta_{l1},\cdots,\theta_{lk} $ are freezed here.

\textbf{LI:} We sample the second batch of trajectories $ \{\tau_{i2}\} $ from $ \mathcal{D}_i^{tr} $. The adapted high level network $ \pi_{\lambda_h} $ will process each state in $ \tau_{i2} $ to get sub-skill category $ \hat{z_{i2t}} = \pi_{\lambda_h}(s_t) $ at each time step, thus we get $ K $ separate data sets for different sub-skills: $ \mathcal{D}_{2k} = \{(s_{i2t}, a_{i2t}) | \hat{z_{i2t}}=k\} $, $ k=1,\cdots,K $. Then we compute the adaptation loss for each sub-skill with the corresponding dataset. In case of continuous action space, we assume that actions belong to Gaussian distributions, so we have:
\begin{equation}
	\label{losslowadapt}
	\mathcal{L}_{BC}(\theta_{lk},\mathcal{D}_{2k})=-\frac{1}{N_k} \sum_{t=1}^{N_k}(a_t - \pi_{\theta_{lk}}(s_t))^2, \end{equation}
where $ N_k $ is the number of state-action pairs in $ \mathcal{D}_{2k} $. Finally we perform gradient descent on sub-skills and get $ \lambda_{lk} = \theta_{lk}-\alpha\nabla_{\theta_{lk}}\mathcal{L}_{BC}(\theta_{lk},\mathcal{D}_{2k}) $. Note $ \pi_{\lambda_h} $ is frozen in this process. 

\textbf{HO:} We sample the third batch of trajectories $ \{\tau_{i3}\} $  from $ \mathcal{D}_i^{tr} $ and get $ \mathcal{L}(\lambda_h,\tau_{i3}) $ as in the HI process. Then we use it to compute the meta-gradient $\nabla_{\theta_h}\mathcal{L}(\lambda_h,\tau_{i3})$ which equals to:
\begin{equation}
	\label{high-update} \nabla_{\lambda_h}\mathcal{L}(\lambda_h,\tau_{i3})|_{\lambda_h = \theta_h - \alpha \nabla_{\theta_h}\mathcal{L}(\theta_h,\tau_{i2})}*\nabla_{\theta_h}\lambda_h.
\end{equation}

\textbf{LO:} We sample $ \tau_{i4} $ and get $ \mathcal{L}(\lambda_{lk},\mathcal{D}_{4k}) $, $ k=1,\cdots,K $ as in the LI process, then we use it to compute the meta-gradient $\nabla_{\theta_{lk}}\mathcal{L}(\lambda_{lk},\mathcal{D}_{4k})$ which equals to:
\begin{equation}
	\label{low-update} \nabla_{\lambda_{lk}}\mathcal{L}(\lambda_{lk},\mathcal{D}_{4k})|_{\lambda_{lk} = \theta_{lk} - \alpha \nabla_{\theta_{lk}}\mathcal{L}(\theta_{lk},\mathcal{D}_{1k})}*\nabla_{\theta_{lk}}\lambda_{lk}.
\end{equation}
Note after the training of $ m $ tasks, we average all meta-gradients from $ m $ tasks and perform gradient descents on the initial parameters \textit{together} to update high-level parameters $ \theta_h' = \theta_h - \beta \sum_{i=1}^{m} \nabla_{\theta_h}\mathcal{L}(\lambda_h,\tau_{i3}) $ and sub-skill policies parameters $ \theta_{lk}' = \theta_{lk}-\beta\sum_{i=1}^{m}\nabla_{\theta_{lk}}\mathcal{L}(\lambda_{lk},\tau_{i4}) $, $ k=1,\cdots,K $, i.e., we do not update them at step \ref{high-update} and \ref{low-update}. This is crucial to ensure convergence.

For testing, although our method needs totally two batches of trajectories for one round of adaptation, in practice we find only using one trajectory to perform HI and LI also works well in new tasks, thus DMIL can satisfy the one-shot imitation learning requirement. Besides the above process, we also add an  auxiliary loss to better drive out meaningful sub-skills to avoid the excessively frequently switching between different sub-skills along with time. Detailed information can be found in \ref{auxiliaryloss}.

\section{Theoretical Analysis}

DMIL is a novel iterative hierarchical meta-learning procedure, and its convergence of DMIL needs to be proved to ensure feasibility. As stated in the above section, what makes DMIL special is that in \textbf{HI} and \textbf{LI}, we update parameters of each module immediately, but in \textbf{LO} and \textbf{HO}, we store the gradients of each part and update them simultaneously. In this section, we show this can make DMIL converge by rewriting both MAML and DMIL as hierarchical variational Bayes problems to establish the equivalence between them, since the convergence of MAML can be proved in \citet{convergenceofmaml}. Proofs of all theorems are in Appendix \ref{proofs}.

\subsection{Hierarchical Variational Bayes Formulation of MAML}

According to \citet{amortizedbayesianmeta}, MAML is a hierarchical variational Bayes inference process. The general meta-learning objective (\ref{metaloss}), which equals to  $\mathcal{L}(\theta,\lambda_1,\cdots,\lambda_{m})$ (we use $ \mathcal{L}_{g} $ for short), can be formulated as follows:
\begin{equation}
	\label{hbayes}
	\begin{split}
		\mathcal{L}_{g} = &
		\log \left[\prod_{i=1}^{m} p\left(\mathcal{D}_{i}|\theta\right)\right] \\ \geq &\sum_{i=1}^{m}\{ \operatorname{KL}(q(\phi_i;\lambda_{i}) \ \| \  p(\phi_i|\mathcal{D}_i,\theta))  \\& \ \ \ +  E_{q(\phi_i;\lambda_i)}[\log p(\mathcal{D}_i,\phi_i|\theta)-\log q(\phi_i;\lambda_i)]\},
	\end{split}
\end{equation}
where $ \phi_i, i=1,\cdots,m $ represent the local latent variables for task $ \mathcal{T}_i $, and $ \lambda_1, \cdots, \lambda_M $ are the variational parameters of the approximate posteriors over $ \phi_1, \cdots, \phi_M $. We denote $ \lambda_{i}$ as $\lambda_{i}(\mathcal{D}_i,\theta) $ and $ p(\phi_i|\mathcal{D}_i,\theta) $ as $ p(\phi_i|\mathcal{D}_i^{tr},\theta) $ to mean that $ \lambda_{i} $ and $ \phi_i $ are determined with prior parameters $ \theta $ and support data $ \mathcal{D}_i^{tr} $. First we need to minimize $  \operatorname{KL}(q(\phi_i;\lambda_{i}) \| p(\phi_i|\mathcal{D}_i^{tr},\theta)) $ w.r.t. $ \lambda_{i} $. According to \ref{proofofeq8}, we have:
\begin{equation}
	\label{findlambda}
	\begin{split}
		\lambda_i(\mathcal{D}_i^{tr},\theta) = \mathop{\arg\max}_{\lambda_i}& E_{q(\phi_i;\lambda_i)}[\log p(\mathcal{D}_i^{tr}|\phi_i)] \\ & -  \operatorname{KL}(q(\phi_i;\lambda_{i}) \| p(\phi_i|\theta)),
	\end{split}
\end{equation}
and we can establish the connection between \ref{findlambda} and the fine-tuning process in MAML by the following Lemma:

\textbf{Lemma 1} In case $ q(\phi_i;\lambda_{i}) $ is a Dirac-delta function and choosing Gaussian prior for $ p(\phi_i|\theta) $, equation \ref{findlambda} equals to the inner-update step of MAML, that is, maximizing $ \log p(\mathcal{D}_i^{tr}) $ w.r.t. $ \lambda_i $ by early-stopping gradient-ascent with choosing $ \mu_\theta $ as initial point:
\begin{equation}
	\label{lemma1}
	\lambda_i(\mathcal{D}_i^{tr};\theta) =\mu_{\theta}+\alpha\nabla_{\theta} \log p\left(\mathcal{D}_i^{tr} | \theta\right)|_{\theta=\mu_{\theta}}.
\end{equation}
Then we need to optimize $ \mathcal{L}(\theta,\lambda_1,\cdots,\lambda_{M}) $ w.r.t. $ \theta $. Since we evaluate $ p(\mathcal{D}_i|\lambda_i(\mathcal{D}_i^{tr},\theta)) $ with only $ \mathcal{D}_i^{val} $, we assume $ p(\mathcal{D}_i|\lambda_i(\mathcal{D}_i^{tr},\theta)) = p(\mathcal{D}_i^{val}|\lambda_i(\mathcal{D}_i^{tr},\theta)) $. We give out the following theorem to establish the connection between the meta-update process and the optimization of $ \mathcal{L}_{g} $:

\textbf{Theorem 1} In case that $ \Sigma_{\theta}\rightarrow 0^+ $, i.e., the uncertainty in the global latent variables $ \theta $ is small, the following equation holds:
\begin{equation}
	\label{findtheta}
	\nabla_\theta \mathcal{L}_{g} = \sum_{i=1}^{M} \nabla_{\lambda_{i}}\log p(\mathcal{D}_i^{val}|\lambda_i)  \nabla_{\theta}\lambda_i(\mathcal{D}_i^{tr},\theta).
\end{equation}
A general EM algorithm will first compute the distribution of latent variables (E-step), then optimize the joint distribution of latent variable and trainable parameters (M-step), and the likelihood of data can be proved to be  monotone increasing to guarantee the convergence since the evidence lower bound of likelihood is monotone increasing. Here $ \phi_i, i=1,\cdots,M $ are the latent variables, and $ \theta $ corresponds to the trainable parameters. Lemma 1 and Theorem 1 correspond to the E-step and M-step respectively. In the following part we establish the equivalence between \ref{lemma1} with \ref{losshighadapt} and \ref{losslowadapt}, and between \ \ref{findtheta} with \ref{high-update} and \ref{low-update} to prove the equivalence between DMIL and MAML.

\subsection{Modeling DMIL with hierarchical variational Bayes framework}

For simplicity, here we only derive in one specific task $ \mathcal{T}_i $, since derivatives of parameters from multi-task can directly add up. We first establish the connection between the maximization  of $ \log p(\mathcal{D}_i^{tr}|\theta_h,\theta_{l1},\cdots,\theta_{lK}) $ with the particular loss functions in DMIL:

\textbf{Theorem 2} In case of $ p(a_t|s_t,\theta_{lk}) \sim \mathcal{N}(\mu_{\theta_{lk}(s_t)}, \sigma^2) $, we have:
\begin{equation}
	\nabla_{\theta_h} \log p(\mathcal{D}_i^{tr}|\theta_h,\theta_{l1},\cdots,\theta_{lK}) = \nabla_{\theta_h} \mathcal{L}_h(\theta_h,\mathcal{D}_i^{tr}),
\end{equation}
and
\begin{equation}
	\label{subskillgradient}
	\nabla_{\theta_{lk}} \log p(\mathcal{D}_i^{tr}|\theta_h,\theta_{l1},\cdots,\theta_{lK}) = \nabla_{\theta_{lk}} \mathcal{L}_{BC}(\theta_{lk},\mathcal{D}_{2k}),
\end{equation}
where $k=1,\cdots,K.$ Note in \ref{subskillgradient}, $ \mathcal{D}_{2k} $ corresponds to data sets determined by the adapted high level network $ \lambda_{h} $, and this connects with \ref{losshighadapt} and \ref{losslowadapt} in DMIL. According to \ref{findlambda}, finding $ \lambda_{i} $ equals to maximize $ \log p\left(\mathcal{D}_i^{tr} | \theta\right) $ in specific conditions, and here in Theorem 2, we prove that maximize $ \log p(\mathcal{D}_i^{tr}|\theta_h,\theta_{l1},\cdots,\theta_{lK}) $ corresponds to \ref{losshighadapt} and \ref{losslowadapt} in DMIL. Thus theorem 2 corresponds to the E-step of DMIL, where we take $ \tau_{i1} $ and $ \tau_{i2} $ as $ \mathcal{D}_i^{tr} $, and optimize $ \mathop{\arg\max}_{\lambda_i} E_{q(\phi_i;\lambda_i)}[\log p(\mathcal{D}_i^{tr}|\phi_i)] -  \operatorname{KL}(q(\phi_i;\lambda_{i}) \| p(\phi_i|\theta)) $ with coordinate descent method, which can be proved to be equal to \ref{lemma1} in \ref{estep}.

For the M-step, we take $ \tau_{i3} $ and $ \tau_{i4} $ as $ \mathcal{D}_i^{val} $. According to Theorem 1, we can take the derivative of $ \lambda_{ih}, \lambda_{il1},\cdots,\lambda_{ilK} $ to maximize the joint distribution of latent variables and trainable parameters to maximize the likelihood of dataset, so we have:
\begin{equation}
	\begin{split}
		&\nabla_{\theta_h,\theta_l}\log p(\mathcal{D}_i^{val}|\lambda_{ih},\lambda_{il})\\=& [\nabla_{\lambda_{ih}}\log p(\mathcal{D}_i^{val}|\lambda_{ih})*\nabla_{\theta_h}\lambda_{ih}(\mathcal{D}_i^{tr},\theta_h),\\ & \ \ \ \ \ \  \nabla_{\lambda_{il}}\log p(\mathcal{D}_i^{val}|\lambda_{il})*\nabla_{\theta_l}\lambda_{il}(\mathcal{D}_i^{tr},\theta_l)]^T
	\end{split}
\end{equation}
where $ \theta_{il} = [\theta_{i1},\cdots,\theta_{iK}]^T $ and $ \lambda_{il} = [\lambda_{i1},\cdots,\lambda_{iK}]^T $. This is exactly the gradients computed in HO and LO steps. Note this computation process can be automatically accomplished with standard deep learning libraries such as PyTorch \cite{pytorch}. To this end, we establish the equivalence between DMIL and MAML, and the convergence of DMIL can be proved.

For a clearer comparison, MAML is an iterative process of $ \theta \rightarrow \lambda \rightarrow \theta'  $, and DMIL is an iterative process of $ \theta_h,\theta_l \rightarrow \lambda_{h},\theta_l \rightarrow \lambda_{h},\lambda_{l} \rightarrow \theta_h',\theta_l' $, where the posterior estimation stages $ \theta_h,\theta_l \rightarrow \lambda_{h},\theta_l \rightarrow \lambda_{h},\lambda_{l} $ has no effect on parameters $ \theta_h,\theta_l $, thus can be divided to two steps as in DMIL. This decoupled fine-tuning fashion is exactly what we need to first adapt the high-level network and then adapt sub-skills. If we end-to-end fine-tune parameters like $ \theta_h,\theta_l \rightarrow \lambda_{h},\lambda_{l} $, sub-skills will receive supervisions from an unadapted high-level network, which may provide incorrect classifications. Different to this, the meta-updating process $ \lambda_{h},\lambda_{l} \rightarrow \theta_h',\theta_l' $ must be done at the same time, since if we update $ \theta_h $ and $ \theta_l $ successively, the later one will receive different derivative (for example, $ \nabla_{\theta_l}\log p(\mathcal{D}_i^{val}|\theta_{ih}',\lambda_{il}) $ ) from derivatives in MAML ($ \nabla_{\theta_l}\log p(\mathcal{D}_i^{val}|\lambda_{ih},\lambda_{il}) $), and the equivalence would not be proved.

\section{Experiments}


In experiments we aim to answer the following questions: (a) Can DMIL successfully transfer the learned hierarchical structure to new tasks with few-shot new task demonstrations? (b) Can DMIL achieve higher performance compared to other few-shot imitation learning methods? (c) What are effects of different parts in DMIL, such as the skill number $ K $, the bi-level meta-learning procedure, and the continuity constraint? Codes are provided in an anonymous repository\footnote{https://anonymous.4open.science/r/DMIL}. Video results are provided in supplementary materials. 


\subsection{Environments and Experiment Setups}

\begin{table*}[t]
	\centering
	\caption{Success rates of different methods on Meta-world environments with $ K=3 $. Each data point comes from 20 random seeds.}
	\resizebox{\textwidth}{!}{
		\begin{tabular}{c|llll|cccc}
			\toprule
			& \multicolumn{4}{c|}{ML10}     & \multicolumn{4}{c}{ML45} \\
			\cmidrule{2-9}          & \multicolumn{2}{c}{Meta-training} & \multicolumn{2}{c|}{Meta-testing} & \multicolumn{2}{c}{Meta-training} & \multicolumn{2}{c}{Meta-testing} \\
			Methods & \multicolumn{1}{c}{1-shot} & \multicolumn{1}{c}{3-shot} & \multicolumn{1}{c}{1-shot} & \multicolumn{1}{c|}{3-shot} & 1-shot & 3-shot & 1-shot & 3-shot \\
			\midrule
			OptionGAIL & \multicolumn{1}{c}{0.455$\pm$0.011} & \multicolumn{1}{c}{\textbf{0.952$\pm$0.016}} & \multicolumn{1}{c}{0.241$\pm$0.042} & \multicolumn{1}{c|}{0.640$\pm$0.025} & 0.506$\pm$0.008 & 0.715$\pm$0.006 & 0.220$\pm$0.013 & 0.481$\pm$0.010 \\
			MIL   & \textbf{0.776$\pm$0.025} & \multicolumn{1}{c}{0.869$\pm$0.029} & \multicolumn{1}{c}{0.361$\pm$0.040} & \multicolumn{1}{c|}{0.689$\pm$0.032} & 0.584$\pm$0.011 & 0.745$\pm$0.017 & 0.205$\pm$0.024 & 0.510$\pm$0.005 \\
			PEMIRL & \multicolumn{1}{c}{0.598$\pm$0.023} & \multicolumn{1}{c}{0.810$\pm$0.007} & \multicolumn{1}{c}{0.162$\pm$0.003} & \multicolumn{1}{c|}{0.256$\pm$0.009} & 0.289$\pm$0.051 & 0.396$\pm$0.024 & 0.105$\pm$0.005 & 0.126$\pm$0.008 \\
			MLSH  & \multicolumn{1}{c}{0.506$\pm$0.134} & \multicolumn{1}{c}{0.725$\pm$0.021} & \multicolumn{1}{c}{0.106$\pm$0.032} & \multicolumn{1}{c|}{0.135$\pm$0.009} & 0.235$\pm$0.093 & 0.295$\pm$0.021 & 0.050$\pm$0.000 & 0.050$\pm$0.000 \\
			DMIL  & 0.775$\pm$0.010 & 0.949$\pm$0.009 & \textbf{0.396$\pm$0.016} & \textbf{0.710$\pm$0.021} & \textbf{0.590$\pm$0.010} & \textbf{0.859$\pm$0.008} & \textbf{0.376$\pm$0.004} & \textbf{0.640$\pm$0.009} \\
			\bottomrule
		\end{tabular}%
	}
	\vskip -0.15in
	\label{tab:successrate}%
\end{table*}%

\begin{table*}[t]
	\centering
	\caption{Cumulative  rewards of different methods on four unseen tasks in Kitchen environment with $ K=4 $. Boldface indicates excluded objects during training.}
	\resizebox{0.9\textwidth}{!}{
		\begin{tabular}{c|ccc}
			\toprule
			Task (Unseen) &  FIST-no-FT & SPiRL & DMIL(ours) \\
			\midrule
			Microwave, Kettle, \textbf{Top Burner}, Light Switch & 2.0 $\pm$ 0.0 & \textbf{2.1 $\pm$ 0.48} & 1.5$\pm$0.48 \\
			\textbf{Microwave}, Bottom Burner, Light Switch, Slide Cabinet & 0.0 $\pm$ 0.0 &  2.3 $\pm$ 0.49 & \textbf{2.35$\pm$0.39} \\
			Microwave, \textbf{Kettle}, Hinge Cabinet, Slide Cabinet & 1.0 $\pm$ 0.0 & 1.9 $\pm$ 0.29 & \textbf{3.15$\pm$0.22} \\
			Microwave, Kettle, Hinge Cabinet, \textbf{Slide Cabinet} & 2.0 $\pm$ 0.0 & \textbf{3.3 $\pm$ 0.38} & 2.95$\pm$0.44 \\
			
			\bottomrule
	\end{tabular}}%
	\vskip -0.1in
	\label{tab:kitchen}%
\end{table*}%

We choose to evaluate DMIL on two representative robot manipulation environments. The first one is  Meta-world benchmark environments \cite{metaworld}, which contains 50 diverse robot manipulation tasks, as shown in \ref{ml45} and \ref{ml10}. We use both ML10 suite (10 meta-training tasks and 5 meta-testing tasks) and ML45 suite (45 meta-training tasks and 5 meta-testing tasks) to evaluate our method, and collect 2K demonstrations for each task. We choose Meta-world since we think a large scale of diverse manipulation tasks can help our method to learn semantic skills. We use the following approaches for comparison in this environment: \textbf{Option-GAIL}: a hierarchical generative adversarial imitation learning method to discover options from unsegmented demonstrations \cite{optiongail}. We use Option-GAIL to evaluate the effect of meta-learning in DMIL. \textbf{MIL}: a transformer-based meta imitation learning method \cite{transformerbasedmeta}. We use MIL to evaluate the effect of the hierarchical structure in DMIL. \textbf{MLSH}: the meta-learning shared hierarchies method \cite{metalearningsharedhierarchies} that relearns the high-level network in every new task. We use MLSH to evaluate the effect of fine-tune (rather than relearn) the high-level network in new tasks. \textbf{PEMIRL}: a contextual meta inverse RL method which transfers the reward function in the new tasks \cite{metairl}. We use PEMIRL to show DMIL can transfer to new tasks that have significantly different reward functions.

The second one is the Kitchen environment of the D4RL benchmark \cite{d4rl}, which contains five different sub-tasks in the same kitchen environment. The accomplishment of each episode requires sequentially completions of four specific sub-tasks, as shown in \ref{kitchenenv}. We use an open demonstration data set \cite{replaypolicylearning} to train our method. During training, we exclude interactions with selected objects and at test-time we provide demonstrations that involve manipulating the excluded object to make them as unseen tasks.  We choose this environment to show DMIL can be used in long-horizon tasks and is robust across different environments. We use two approaches for comparison in this experiment: \textbf{SPiRL}: an extension of the skill extraction methods to imitation learning over skill space \cite{spirl}; \textbf{FIST}: an algorithm that extracts skills from offline data with an inverse  dynamics model and a distance function \cite{fist}.

We use fully-connected neuron networks for both the high-level network and sub-skills. Detailed descriptions on the environment setup, demonstration collection procedure, hyper-parameters setting, training process, and descriptions of different methods can be found in appendix \ref{appendix:experiment}.

\subsection{Results}


\begin{figure*}[t]
	\centering
	\includegraphics[width=0.95\linewidth]{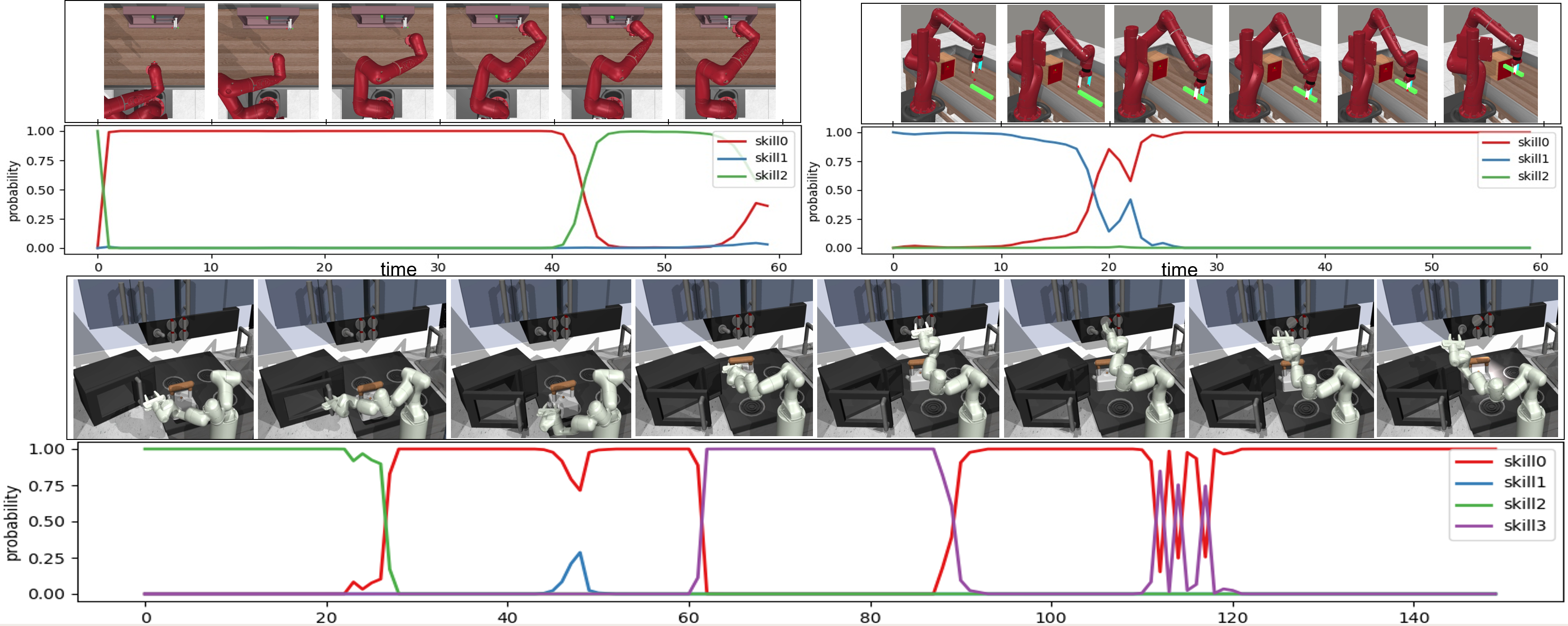}
	\caption{The iterative meta-learning process of DMIL at each iteration. Left: the supervision of high-level network (sub-skill category) comes from the most accurate sub-skill. Right: the sub-skill updated at current step is determined by the fine-tuned high-level network.\label{skillalongtime}}
\end{figure*}

\ref{tab:successrate} shows success rates of different methods in ML10 and ML45 suites with sub-skill number $ K=3 $. We perform 1-shot and 3-shot experiments respectively to show the progressive few-shot performance of different methods. DMIL achieves the best results in ML10 testing suite and ML45 training and testing suites. This shows the superiority of our method for transferring across a large scale of manipulation tasks. OptionGAIL achieves high success rates in both ML10 and ML45 training suites. These results come from their hierarchical structures that have adequate capacity to fit potential multi-modal behaviors in multi-task demonstrations. MIL achieves comparable results for all meta-testing tasks but is  worse than DMIL. This shows the necessity of meta-learning processes. Compared to them,  PEMIRL and MLSH are mediocre among all suites. This comes from that the reward functions across different tasks are difficult to transfer with few shot demonstrations, and the relearned high-level network of MLSH damages previously learned knowledge. We also illustrate t-sne results of these methods in \ref{tsne} to further analyze them in appendix \ref{tsnesection}.

\ref{tab:kitchen} shows the rewards of different methods on four unseen tasks in the Kitchen environment. \textit{FIST-no-FT} refers to a variant of FIST that does not use future state conditioning, which makes the comparison fairer. DMIL achieves higher rewards on two out of four tasks and comparable results on the other two tasks, which exhibits the effectiveness of the bi-level meta-training procedure. The poor performance of DMIL on the first task may come from the choice of skill number $ K $ or from low-quality demonstrations. We perform ablation studies on $ K $ in the next section.

\ref{skillalongtime} shows curves of sub-skill probabilities along time of two tasks \textit{window-close} and \textit{peg-insert-side} of Meta-world and the \textit{microwave-kettle-top burner-light} task in Kitchen environment. We can see the activation of sub-skills shows a strong correlation to different stages of tasks. In first two tasks, $ \pi_{\theta_{l0}} $ activates when the robot is closing to something, $ \pi_{\theta_{l1}} $ activates when the robot is picking up something, and $ \pi_{\theta_{l2}} $  activates when the robot is manipulating something. In the third task, $ \pi_{\theta_{l2}} $ activates when the robot is manipulating the microwave, $ \pi_{\theta_{l0}} $ activates when the robot is manipulating the kettle or the light switch, and $ \pi_{\theta_{l3}} $ activates when the robot is manipulating the burner switch. This shows that DMIL has the ability to learn semantic sub-skills from unsegmented multi-task demonstrations.

\subsection{Ablation Studies}

\begin{table}
	\centering
	\caption{Success rates of different sub-skill number in Meta-world environments.\label{tab:ablationofK}} 

	\resizebox{1\linewidth}{!}{
		\begin{tabular}{c|cccc|cccc}
			\toprule
			& \multicolumn{4}{c|}{ML10}     & \multicolumn{4}{c}{ML45} \\
			\cmidrule{2-9}          & \multicolumn{2}{c}{Meta-training} & \multicolumn{2}{c|}{Meta-testing} & \multicolumn{2}{c}{Meta-training} & \multicolumn{2}{c}{Meta-testing} \\
			K     & 1-shot & 3-shot & 1-shot & 3-shot & 1-shot & 3-shot & 1-shot & 3-shot \\
			\midrule
			2     & 0.76  & 0.955 & 0.32  & \textbf{0.72} & 0.563 & 0.818 & \textbf{0.44} & \textbf{0.67} \\
			3     & 0.775 & 0.949 & 0.396 & 0.71  & 0.59  & 0.859 & 0.376 & 0.64 \\
			5     & 0.795 & 0.94  & \textbf{0.52} & 0.57  & 0.713 & 0.92  & 0.21  & 0.48 \\
			10    & \textbf{0.8} & \textbf{0.975} & 0.38  & 0.62  & \textbf{0.736} & \textbf{0.931} & 0.34  & 0.64 \\
			\bottomrule
	\end{tabular}}
\end{table}

In this section we perform ablation studies on different components of DMIL to provide effects of different parts. Due to limited space, we put ablations on fine-tuning steps, bi-level meta-learning processes, continuity constraint and hard/soft EM choices in appendix \ref{moreresults}.

\textbf{Effect of different skill number $K$}:  \ref{tab:ablationofK} shows the effect of different sub-skill number $ K $ in Meta-world experiments. We can see that a larger $ K $ can lead to higher success rates on meta-training tasks, but a smaller $ K $ can lead to better results on meta-testing tasks. This tells us that an excessive number of sub-skills may result in over-fitting on training data, and a smaller $ K $ can play the role of regularization. In Kitchen experiments, we can see similar phenomenons in \ref{tab:ablationofkofkitchen}.  It is worth noting in both environments, we did not encounter collapse problems, i.e., every sub-skill gets well-trained even when $K=8$ in kitchen environment or $K=10$ in Meta-world environments. This may come from that, in the meta-training stage, adding more sub-skills can help the whole structure get lower loss, thus DMIL will use all of them for training. However, in our supplementary videos, we can see sub-skills trained with a large $K$ (for instance, $K=10$ in Meta-world environments) are not as semantic as sub-skills trained by a small $K$ (for instance, $K=3$ in Meta-world environments) during the execution of a task. 


\section{Discussion and Future Work}


In this work, we propose DMIL to meta-learn a hierarchical structure from unsegmented multi-task demonstrations to endow it with fast adaptation ability to transfer to new tasks. We theoretically proved its convergence by reframing MAML and DMIL as hierarchical Bayes inference processes to get their equivalence. Empirically, we successfully acquire transferable hierarchical structures in both Meta-world and Kitchen Environments. Generally speaking, DMIL can be regarded from two different perspectives: a hierarchical extension of MAML to better perform in multi-modal task distributions, or a meta-learning enhancement of skill discovery methods to make learned skills  quickly transferable to new tasks.

The limitations of DMIL come from several aspects, and future works can seek meaningful extensions in these perspectives. Firstly, DMIL models all tasks as bi-level structures. However, in real-world situations, tasks may be multi-level structures. One can extend DMIL to multi-level hierarchical structures like done in recent works \cite{hierarchicalmultitask}. Secondly, DMIL does not capture temporal information in demonstrations. Future state conditioning in \citet{fist} seems an effective tool to improve few-shot imitation learning performance in long-horizon tasks such as in the Kitchen environments. Future works can employ temporal modules such as transformer \cite{transformer} as the high-level network of DMIL to improve its performance. 

\bibliography{iclr2022_conference}
\bibliographystyle{iclr2022_conference}

\newpage
\appendix
\section{Algorithm}

\begin{algorithm}[h]
	\caption{Dual Meta Imitation Learning}
	\label{alg}
	\begin{algorithmic}
		\STATE {\bfseries Require:} task distribution $p(\mathcal{T}) $, multi-task demonstrations $ \{\mathcal{D}_i\}, i=1,\cdots,m $, initial parameters of high-level network $ \theta_h $ and sub-skill policies $ \theta_{l1}, \cdots, \theta_{lK} $, inner and outer learning rate $\alpha, \beta$.
		
		\STATE {\bfseries while} not done {\bfseries do}
		
		\STATE \quad Sample batch of tasks $ \mathcal{T}_i \sim p(\mathcal{T}) $
		
		\STATE \quad {\bfseries for all $ \mathcal{T}_i $ do}
		
		\STATE \quad \quad Sample $ \{\tau_{i1}\}, \{\tau_{i2}\}, \{\tau_{i3}\}, \{\tau_{i4}\}$ from $ \{\mathcal{D}_i\} $
		
		\STATE \quad \quad Evaluate $ \nabla_{\theta_h}\mathcal{L}_h(\theta_h,\tau_{i1}) $ according to \ref{losshighadapt} and $ \tau_{i1} $
		
		\STATE \quad \quad Compute adapted parameters of high-level network: $ \lambda_h = \theta_h-\alpha\nabla_{\theta_h}\mathcal{L}_h(\theta_h,\tau_{i1}) $
		
		\STATE \quad \quad Evaluate $ \nabla_{\theta_{lk}}\mathcal{L}_{BC}(\theta_{lk},\mathcal{D}_{2k}) $ according to \ref{losslowadapt} and $ \tau_{i2} $, $ k=1,\cdots,K $
		
		\STATE \quad \quad Compute adapted parameters of sub-skills: $ \lambda_{lk} = \theta_{lk}-\alpha\nabla_{\theta_{lk}}\mathcal{L}_{BC}(\theta_{lk},\mathcal{D}_{2k}) $, $ k=1,\cdots,K $

		\STATE \quad \quad Evaluate $ \nabla_{\theta_h}\mathcal{L}_{\mathcal{T}_i}(\lambda_h,\tau_{i3}) $ and $ \nabla_{\theta_{lk}}\mathcal{L}_{\mathcal{T}_i}(\lambda_{lk},\mathcal{D}_{4k}) $, $ k=1,\cdots,K $

		\STATE \quad {\bfseries end for}
		
		\STATE \quad Update $ \theta_h \leftarrow \theta_h-\beta\nabla_{\theta_h}\sum_{\mathcal{T}_i\sim p(\mathcal{T})}\mathcal{L}_{\mathcal{T}_i}(\lambda_h,\tau_{i3}) $
		
		\STATE \quad Update $ \theta_{lk} \leftarrow \theta_{lk}-\beta\nabla_{\theta_{lk}}\sum_{\mathcal{T}_i\sim p(\mathcal{T})}\mathcal{L}_{\mathcal{T}_i}(\lambda_{lk},\mathcal{D}_{4k}) $, $ k=1,\cdots,K $
		
		\STATE {\bfseries end while}
		
	\end{algorithmic}
	
\end{algorithm}

\section{Auxiliary Loss}
\label{auxiliaryloss}

We adopt an auxiliary loss for DMIL to better drive out meaningful sub-skills by punishing excessive switching of sub-skills along the trajectory. This comes from an intuitive idea: each sub-skill should be a temporal-extended macro-action, and the high-level policy only needs to switch to different skills few times along a task, as the same idea of \textit{macro-action} in MLSH \cite{metalearningsharedhierarchies}. We denote:
\begin{equation}
	\label{sign}
	sign(x) =\left\{
	\begin{array}{lr}
		1, \ \text{if} \  x = True &  \\
		0, \ \text{if} x = False &  
	\end{array}
	\right.
	,
\end{equation}
and the auxiliary loss is:
\begin{equation}\label{auloss}
	\mathcal{L}_{aux}(\tau) = \sum_{t=0}^{T-1}sign(\hat{z}_{t+1}\not=\hat{z}_t)\ / \ len(\tau).
\end{equation}

Although this operation seems discrete, in practice we can use the operations in modern deep learning framework such as PyTorch \cite{pytorch} to make it differentiable. We add this loss function to the $ \mathcal{L}_h(\theta_h,\tau_{i1}) $ and $ \nabla_{\theta_h}\mathcal{L}(\lambda_h,\tau_{i3}) $ with a coefficient $ \lambda = 1 $. We also perform ablation studies of $ \mathcal{L}_{aux} $ and results are in   \ref{tab:ablationofconti}.

\section{Proofs}
\label{proofs}

\subsection{Proof of Lemma 1}

\textbf{Lemma 1} In case $ q(\phi_i;\lambda_{i}) $ is a Dirac-delta function and choosing Gaussian prior for $ p(\phi_i|\theta) $, equation \ref{findlambda} equals to the inner-update step of MAML, that is, maximizing $ \log p(\mathcal{D}_i^{tr}) $ w.r.t. $ \lambda_i $ by early-stopping gradient-ascent with choosing $ \mu_\theta $ as initial point:
\begin{equation}
	\lambda_i(\mathcal{D}_i^{tr};\theta) =\mu_{\theta}+\alpha\nabla_{\theta} \log p\left(\mathcal{D}_i^{tr} | \theta\right)|_{\theta=\mu_{\theta}}.
\end{equation}
\textit{Proof:} in case of the conditions of Lemma 1, we have:
\begin{equation}
	\label{lemma1second}
	\lambda_i(\mathcal{D}_i^{tr};\theta) = \mathop{\arg\max}_{\lambda_i}[\log p(\mathcal{D}_i^{tr}|\mu_{\lambda_{i}})-\left\|\mu_{\lambda_{i}}-\mu_{\theta}\right\|^{2} /2 \Sigma_{\theta}^{2}],
\end{equation} 
As stated in \cite{recasting}, firstly in the case of linear models, early stopping of an iterative gradient descent process of $ \lambda $ equals to the maximum posterior estimation (MAP) \cite{earlylinear}. In our case the posterior distribution refers to $ q(\phi_i|\lambda_{i})  $, and MAML is a Bayes process to find the MAP estimate as the point estimate of $ \lambda_i(\mathcal{D}_i^{tr};\theta) $. In the nonlinear case, this point estimate is not necessarily the global mode of the posterior, and we can refer to \cite{earlystopping} for another implicit posterior distribution over $ \phi_i $ and making the early stopping procedure of MAML acting as priors to get the similar result.

\subsection{Proof of Equation \ref{findlambda}}
\label{proofofeq8}

Equation \ref{findlambda} can be written as:
\begin{equation}
	\begin{split}
		\lambda_i(\mathcal{D}_i^{tr},\theta) & =  \mathop{\arg\min}_{\lambda_i} \operatorname{KL}(q(\phi_i;\lambda_{i}) \| p(\phi_i|\mathcal{D}_i^{tr},\theta)) \\ & =  \mathop{\arg\max}_{\lambda_i} E_{q(\phi_i;\lambda_i)}[\log p(\phi_i|\mathcal{D}_i^{tr},\theta) - \log q(\phi_i;\lambda_{i})] \\ & = \mathop{\arg\max}_{\lambda_i} E_{q(\phi_i;\lambda_i)}[\log p(\mathcal{D}_i^{tr}|\phi_i)] -  \operatorname{KL}(q(\phi_i;\lambda_{i}) \| p(\phi_i|\theta)),
	\end{split}
\end{equation}
where in MAML we assume $ p(\mathcal{D}_i^{tr}|\phi_i) = p(\mathcal{D}_i^{tr}|\phi_i,\theta) $, and use the joint distribution $ p(\mathcal{D}_i^{tr},\phi_i|\theta) $ to replace $ p(\phi_i|\mathcal{D}_i^{tr},\theta) $ since we assume that $ p(\mathcal{D}_i^{tr}) $ subjects to uniform distribution. Thus \ref{findlambda} can be proved.

\subsection{Proof of Theorem 1}

\textbf{Theorem 1} In case that $ \Sigma_{\theta}\rightarrow 0^+ $, i.e., the uncertainty in the global latent variables $ \theta $ is small, the following equation holds:
\begin{equation}
	\nabla_\theta \mathcal{L}(\theta,\lambda_1,\cdots,\lambda_{M}) = \sum_{i=1}^{M} \nabla_{\lambda_{i}}\log p(\mathcal{D}_i^{val}|\lambda_i) * \nabla_{\theta}\lambda_i(\mathcal{D}_i^{tr},\theta).
\end{equation}
\textit{Proof:} 
\begin{equation}
	\label{metaupdate}
	\begin{split}
		\nabla_\theta \mathcal{L}(\theta,\lambda_1,\cdots,\lambda_{M}) &\approx \sum_{i=1}^{M}\{ \nabla_\theta E_{q(\phi_i;\lambda_i)}[\log p(\mathcal{D}_i,\phi_i|\theta)-\log q(\phi_i;\lambda_i)]\}
		\\&=\sum_{i=1}^{M} \nabla_{\theta}[\log p(\mathcal{D}_i^{val}|\lambda_i(\mathcal{D}_i^{tr},\theta))-\log p(\lambda_i(\mathcal{D}_i^{tr},\theta)|\theta)]\\& \approx \sum_{i=1}^{M} \nabla_{\theta}\log p(\mathcal{D}_i^{val}|\lambda_i(\mathcal{D}_i^{tr},\theta))\\&
		=\sum_{i=1}^{M} \nabla_{\lambda_{i}}\log p(\mathcal{D}_i^{val}|\lambda_i) * \nabla_{\theta}\lambda_i(\mathcal{D}_i^{tr},\theta)
	\end{split}
\end{equation}

where the first approximate equal holds because the VI approximation error is small enough, and the second approximate equal holds because that in case $ \Sigma_{\theta}\rightarrow 0^+ $ and assuming $ \lambda_i $ be a neuron network, $ \log p(\lambda_i(\mathcal{D}_i^{tr},\theta)|\theta) \approx 0 $ holds almost everywhere, so $ \nabla_\theta\log p(\lambda_i(\mathcal{D}_i^{tr},\theta)|\theta) \approx 0$. Note the condition of theorem 1 is usually satisfied since we are using MAML, and the initial parameters $ \theta $ are assumed to be deterministic. 

From another perspective, the right side of above equation is the widely used meta-gradient in MAML, and it is equal to $ \frac{1}{m} \sum_{i=1}^{m} \left(I-\alpha \nabla^{2}_\theta \mathcal{L}_{BC}\left(\theta, \mathcal{D}_{i}^{tr}\right)\right) *\nabla_{\lambda_i} \mathcal{L}_{BC}\left(\theta-\alpha \nabla_\theta \mathcal{L}_{BC}\left(\theta, \mathcal{D}_{i}^{tr}\right), \mathcal{D}_{i}^{val}\right) $, which is proved to be converged by \cite{convergenceofmaml}. 

\subsection{Proof of Theorem 2}

\textbf{Theorem 2} In case of $ p(a_t|s_t,\theta_{lk}) \sim \mathcal{N}(\mu_{\theta_{lk}(s_t)}, \sigma^2) $, we have:
\begin{equation}
	\label{equal2}
	\nabla_{\theta_h} \log p(\mathcal{D}_i^{tr}|\theta_h,\theta_{l1},\cdots,\theta_{lK}) = \nabla_{\theta_h} \mathcal{L}_h(\theta_h,\mathcal{D}_i^{tr}),
\end{equation}
and
\begin{equation}
	\label{equal1}
	\nabla_{\theta_{lk}} \log p(\mathcal{D}_i^{tr}|\theta_h,\theta_{l1},\cdots,\theta_{lK}) = \nabla_{\theta_{lk}} \mathcal{L}_{BC}(\theta_{lk},\mathcal{D}_{2k}), k=1,\cdots,K.
\end{equation}
\textit{Proof:}
since $ p(\mathcal{D}_i^{tr}|\theta_h,\theta_{l1},\cdots,\theta_{lK}) =\prod_{t=1}^{N}p(a_t|s_t,\theta_h,\theta_{l1},\cdots,\theta_{lK})p(s_t|\theta_h,\theta_{l1},\cdots,\theta_{lK})$ and the second term is independent of $ \theta $, we consider the first conditional probability:
\begin{equation}
	\label{probresolve}
	p(a_t|s_t,\theta_h,\theta_{l1},\cdots,\theta_{lK})=\sum_{k=1}^{K}p(z_k|s_t,\theta_h)p(a_t|s_t,\theta_{lk}).
\end{equation}

In the HI step, $ p(a_t|s_t,\theta_{lk}) $ is fixed, thus \ref{probresolve} becomes a convex optimization problem:
\begin{equation}
	\label{convex}
	\begin{split}
		\mathop{\max}_{\theta_h}\ \sum_{k=1}^{K}&p(z_k|s_t,\theta_h)p(a_t|s_t,\theta
	\end{split}
\end{equation}
The solution of this problem is 
$ \lambda_h^*$ which satisfies $ p(z_k|s_t,a_t,\lambda_h^*)=1 $, $ k=\mathop{\arg\max}_{k}p(a_t|s_t,\theta_{lk}) $. This means that $ \pi_{\theta_h} $ needs to predict the sub-skill category at time step $ t $ as $ k $, in which case $ \pi_{\theta_{lk}} $ can maximize $ p(a_t|s_t,\theta_{lk}) $. In case we choose $ \pi_{\theta_h} $ to be a classifier that employs a Softmax layer at the end, minimizing the cross entropy loss \ref{losshighadapt} equals to maximize \ref{convex}, thus \ref{equal2} can be proved.

In the LI step, $ p(z_k|s_t,\lambda_h) $ is fixed, and the data sets for optimizing $ \theta_{l1},\cdots,\theta_{lK} $ are also fixed as $ \mathcal{D}_{2k} = \{(s_{ijt}, a_{ijt}) | \hat{z_{i2t}}=k\}_{t=1}^{N_k} $. Thus we need to maximize each $ p(a_t|s_t,\theta_{lk}) $ with $ \mathcal{D}_{2k} $. In case of $ p(a_t|s_t,\theta_{lk}) \sim \mathcal{N}(\mu_{\theta_{lk}(s_t)}, \sigma^2) \propto \exp [-\frac{(a_t-\pi_{\theta_{lk}}(s_t))^2}{2\sigma^2}] $, we have $ \mathop{\max}_{\theta_{lk}}p(a_t|s_t,\theta_{lk}) \Leftrightarrow \mathop{\min}_{\theta_{lk}}(a_t-\pi_{\theta_{lk}}(s_t))^2$, which leads to the loss function \ref{losslowadapt}, thus \ref{equal1} can be proved, which finishes the prove of Theorem 2.

\subsection{Proof of the E-step of DMIL}
\label{estep}

According to Theorem 1, we aim to maximize \ref{lemma1second} w.r.t $ \lambda_i $ from the initial point $ \theta_i $ with coordinate gradient ascent. We here need to prove that in DMIL, we could also achieve the global maximum point of $ \lambda_i $ as in MAML. We first give out the following Lemma:

\textbf{Lemma 2} Let $ x $ be the solution found by coordinate gradient descent of $ f(x) $. Let $ x_i, i=1,\cdots,n $ be the $ n $ coordinate directions used in the optimization process. If $ f(x) $ can be decomposed as:
\begin{equation}
	f(x) = g(x) + \sum_{i=1}^{n}h_i(x),
\end{equation}
where $ g(x) $ is a differentiable convex function, and each $ h_i(x) $ is a convex function of the coordinate direction $ x_i $, then $ x $ is the global minimum of $ f(x) $.

\textit{Proof:} Let $ y $ be another arbitrary point, we have:
\begin{equation}
	\begin{split}
		f(y) - f(x) &= g(y) + h(y) - (g(x)+h(x))\\ &\geq \nabla_x g(x)^T(y-x)+\sum_{i=1}^{n}h_i(y_i)-h_i(x_i)\\& = \sum_{i=1}^{n} (\nabla_i g(x)(y_i-x_i) + h_i(y_i)- h_i(x_i)) \\& \geq 0.
	\end{split}
\end{equation}

Now let's consider our problem. Consider 
\begin{equation}
	\begin{split}
		\log p(\mathcal{D}_i^{tr}|\theta_{ih},\theta_{il}) &= \log \sum_{t=1}^{T}p(a_t|s_t,\theta_{ih},\theta_{il})p(s_t) \\&= \log \sum_{t=1}^{T} p(s_t)\sum_{k=1}^{K}p(z_k|s_t,a_t,\theta_{ih})p(a_t|s_t,\theta_{il})\\& \geq \sum_{t=1}^{T}[\log p(s_t)+\log \sum_{k=1}^{K}p(z_k|s_t,a_t,\theta_{ih})p(a_t|s_t,\theta_{il})]\\& \geq \sum_{t=1}^{T}[\log p(s_t)+ \sum_{k=1}^{K}\log p(z_k|s_t,a_t,\theta_{ih}) p(a_t|s_t,\theta_{il})]\\&=\sum_{t=1}^{T}[\log p(s_t)+ \sum_{k=1}^{K}\log p(z_k|s_t,a_t,\theta_{ih}) + \sum_{k=1}^{K}\log  p(a_t|s_t,\theta_{il})].
	\end{split}
\end{equation}

In our case, two coordinate directions are $ \theta_{ih} $ and $ \theta_{il} $. Let's consider the terms inside the brackets. According to Lemma 2, we can think $ \log p(s_t) $ as $ g(x) $(here it equals to constant), $ \sum_{k=1}^{K}\log p(z_k|s_t,a_t,\theta_{ih}) $ as $ h_1(x) $ and $ \sum_{k=1}^{K}\log  p(a_t|s_t,\theta_{il}) $ as $ h_2(x) $. Thus the optimum can be proved.

\section{Additional Ablation Studies}
\label{moreresults}

\subsection{Effects of the Bi-level Meta-learning Process}

We use two variants of DMIL to see the effectiveness of bi-level meta-earning process. \textbf{DMIL-High}: a variant that only meta-learns the high-level network, and \textbf{DMIL-Low}: a variant that only meta-learns sub-skills. We use Option-GAIL as a comparison that does not meta-learn any level of the hierarchical structure.

\ref{tab:ablationofbilevel} shows the results of this ablation study. DMIL-High achieves close results with OptionGAIL in all meta-training suites and better results in all meta-testing suites, but worse than DMIL in all cases. This shows that meta-learning the high-level network can help the hierarchical structure adapt to new tasks, but only transferring the high-level network is not enough for accomplish all kinds of new tasks. DMIL-Low achieves poor results in all suites except in ML10 meta-training suites. This shows that transferring the high-level network is necessary when training on a large scale of tasks or testing in new tasks.

\begin{table}[h]
	\centering
	\caption{Success rates of DMIL-High, DMIL-Low, DMIL and OptionGAIL on Meta-world environments with $ K=3 $. Each data point comes from the success rate of 20 tests.}
	\resizebox{\textwidth}{!}{
		\begin{tabular}{c|llll|cccc}
			\toprule
			& \multicolumn{4}{c|}{ML10}     & \multicolumn{4}{c}{ML45} \\
			\cmidrule{2-9}          & \multicolumn{2}{c}{Meta-training} & \multicolumn{2}{c|}{Meta-testing} & \multicolumn{2}{c}{Meta-training} & \multicolumn{2}{c}{Meta-testing} \\
			Methods & \multicolumn{1}{c}{1-shot} & \multicolumn{1}{c}{3-shot} & \multicolumn{1}{c}{1-shot} & \multicolumn{1}{c|}{3-shot} & 1-shot & 3-shot & 1-shot & 3-shot \\
			\midrule
			OptionGAIL & \multicolumn{1}{c}{0.755$\pm$0.011} & \multicolumn{1}{c}{\textbf{0.952$\pm$0.016}} & \multicolumn{1}{c}{0.241$\pm$0.042} & \multicolumn{1}{c|}{0.640$\pm$0.025} & 0.506$\pm$0.008 & 0.715$\pm$0.006 & 0.220$\pm$0.013 & 0.481$\pm$0.010 \\
			\midrule
			DMIL-High & \multicolumn{1}{c}{0.634$\pm$0.001} & \multicolumn{1}{c}{0.914$\pm$0.011} & \multicolumn{1}{c}{0.298$\pm$0.012} & \multicolumn{1}{c|}{0.670$\pm$0.015} & 0.495$\pm$0.007 & 0.735$\pm$0.006 & 0.280$\pm$0.016 & 0.551$\pm$0.011 \\
			DMIL-Low & \multicolumn{1}{c}{0.746$\pm$0.011} & \multicolumn{1}{c}{0.943$\pm$0.006} & \multicolumn{1}{c}{0.291$\pm$0.024} & \multicolumn{1}{c|}{0.666$\pm$0.021} & 0.511$\pm$0.005 & 0.765$\pm$0.009 & 0.266$\pm$0.010 & 0.492$\pm$0.014 \\
			DMIL  & \textbf{0.775$\pm$0.010} & 0.949$\pm$0.009 & \textbf{0.396$\pm$0.016} & \textbf{0.710$\pm$0.021} & \textbf{0.590$\pm$0.010} & \textbf{0.859$\pm$0.008} & \textbf{0.376$\pm$0.004} & \textbf{0.640$\pm$0.009} \\
			\bottomrule
	\end{tabular}}%
	\label{tab:ablationofbilevel}%
\end{table}%

\begin{figure}[t]
	\centering  
	\subfigure[T-sne results of demonstrations and sub-skill categories of several hierarchical models for meta-testing task \textit{hand-insert}.]{
		\label{tsne}
		\includegraphics[width=0.62\textwidth]{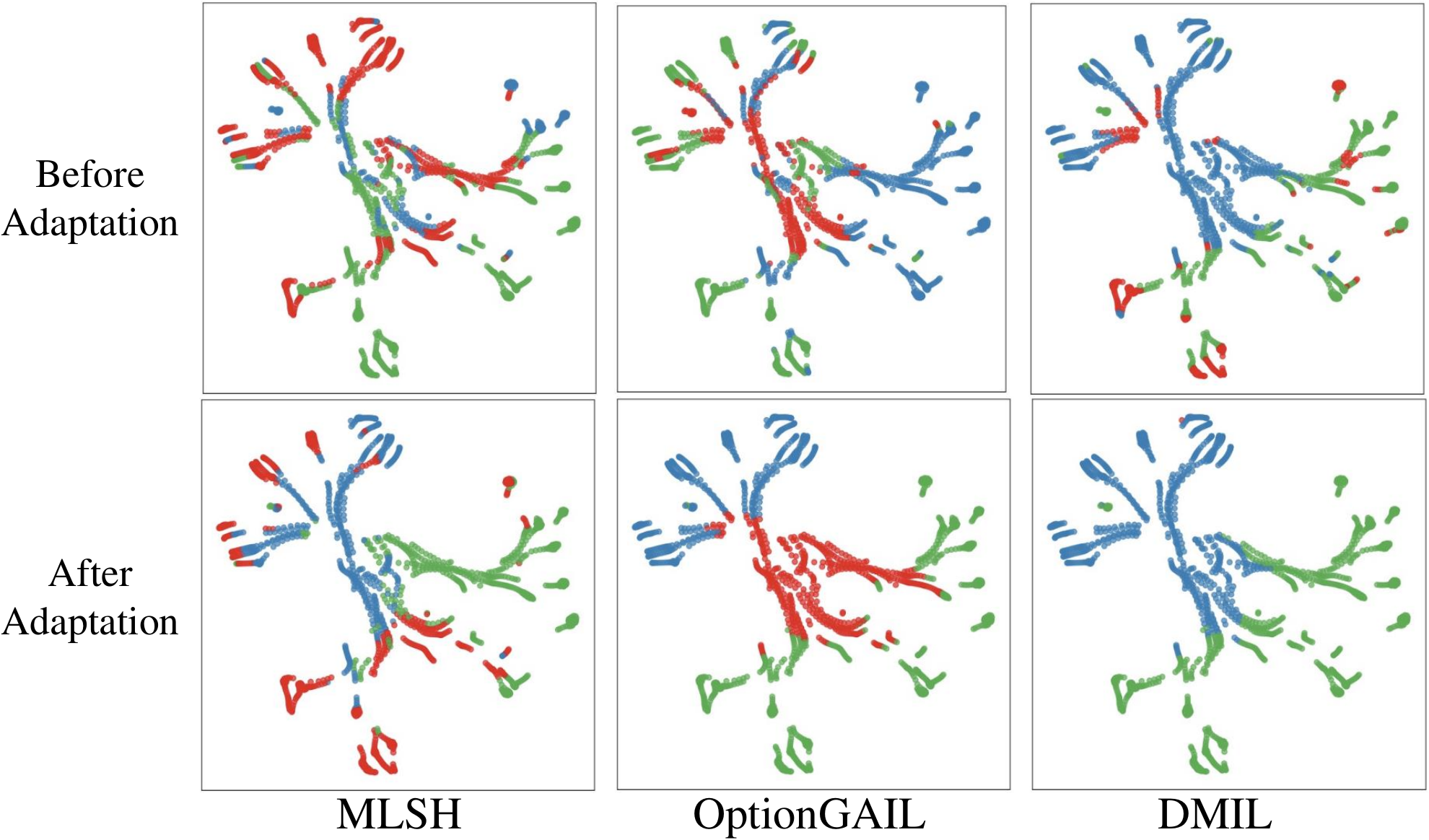}}
	\subfigure[Sub-skill categories of task \textit{push-around-wall} for ablation study.]{
		\label{fig:abalation}
		\includegraphics[width=0.35\textwidth]{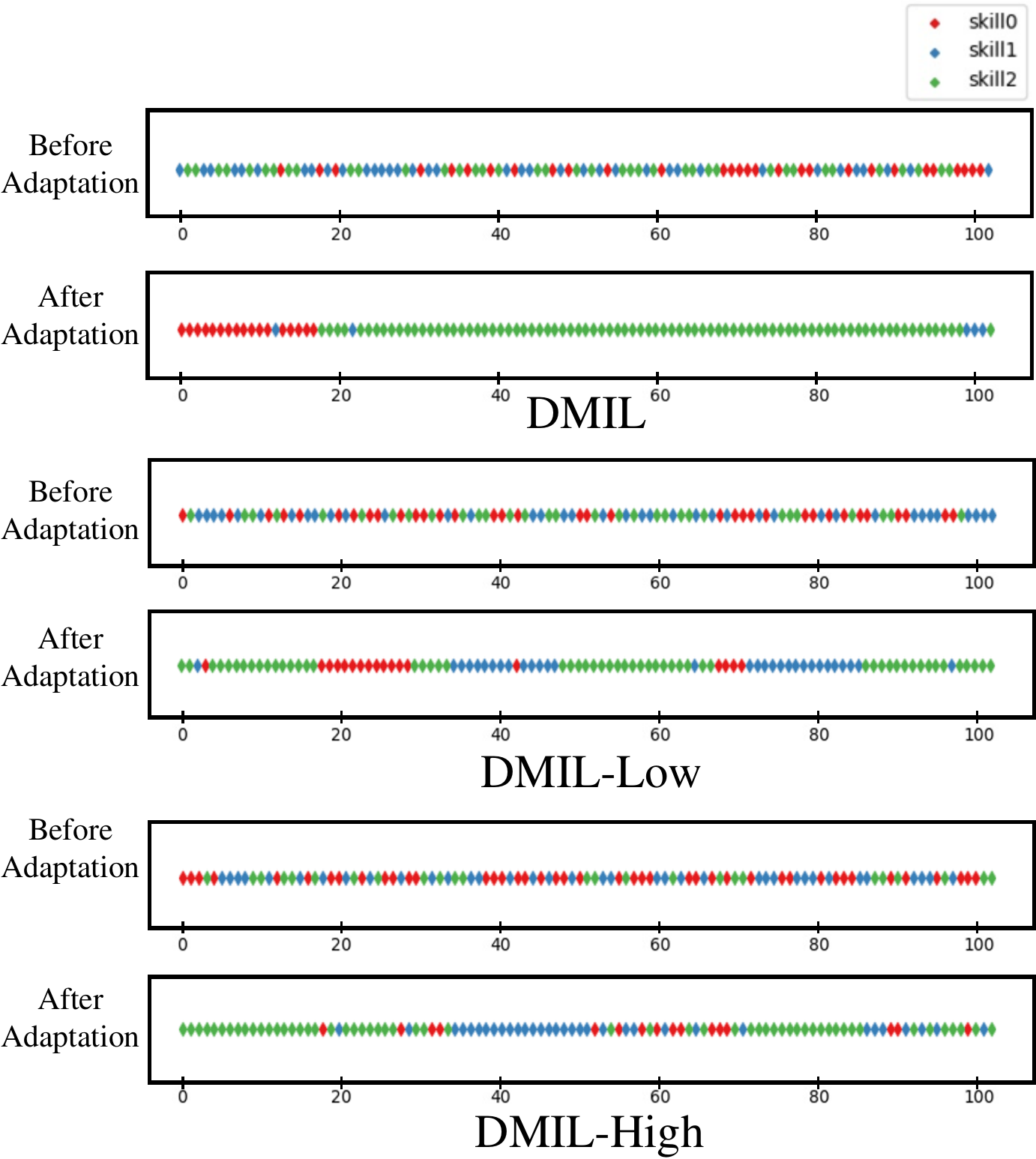}}
	\caption{T-sne results and ablation studies about the bi-level meta-learning process.}
	\label{fig:pics}
\end{figure}

For better understand the effect of bi-level meta-learning process, we perform ablation study for three DMIL variants in an manually-designed new task \textit{push-around-wall} (\ref{pusharoundwall}). In this task, the robot needs to grasp a cube and circle it around the wall. This is a brand new skill that is not in the meta-world suite.  The accomplishment of this new task requires quickly adapting abilities of both the high-level network and sub-skills. We sample two demonstrations and use the first one as few-shot data, and illustrate the sub-skill categories of the second  demonstration given by the high-level network at  \ref{fig:abalation}. Before adaptation, all variants give out approximately random results. However, after one-shot adaptation, DMIL classifies almost every state into sub-skill 0 and sub-skill 2, which indicates these two sub-skills in DMIL have been adapted to the new task, and the high-level network has also been adapted with the supervision from adapted sub-skills. Compared to DMIL, DMIL-Low still can not give out reasonable results after adaptation, since its high-level network lacks the ability to quickly transfer to new tasks. Instead, DMIL-High gives out plausible results after adaptation. This shows the high-level network has adapted to the new task according to the supervision from adapted sub-skills, but no sub-skill can dominate for a long time period since all sub-skills lack the quickly adaptation ability.

\subsection{T-sne Results of Different Methods}
\label{tsnesection}

For comparison of different methods, we illustrate the t-sne results of states of each sub-skill in an
ML45 meta-testing task hand-insert in \ref{tsne}. We use 3 demonstrations for adaptation, and draw the t-sne results on another 16 demonstrations. This task is a meta-testing task, so no method has
ever been trained on this task before.

MLSH shows almost random clustering results no matter before and after adaptation, since its high-level network is relearned in new tasks. OptionGAIL clusters to three sub-skills after adaptation. Compared to them, DMIL clusters the data to only two sub-skills after adaptation. We believe fewer categories reflect  more meaningful sub-skills are developed in DMIL.

\subsection{Effects of Sub-skill Number K in The Kitchen Environment}

We perform ablation studies of sub-skill number $K$ on the Kitchen environments and choose $K=2,4,8$ respectively.   \ref{tab:ablationofkofkitchen} shows the results. We can see that a smaller number of sub-skills can achieve better results on such four unseen results that a large number of sub-skills. This may indicate that the sub-skill number $K$ can work as a 'bottleneck' like the middle layer in an auto-encoder.

\begin{table}[h]
	\centering
	\caption{Ablations of sub-skill number $K$ in Kitchen environments.}
	\begin{tabular}{c|ccc}
		\toprule
		Task (Unseen) &  K=2  & K=4   & K=8 \\
		\midrule
		Microwave, Kettle, Top Burner, Light Switch & \textbf{1.9$\pm$0.43} & 1.5$\pm$0.48 & 1.7$\pm$0.22 \\
		Microwave, Bottom Burner, Light Switch, Slide Cabinet & 2.15$\pm$0.19 & \textbf{2.35$\pm$0.39} & 2.0$\pm$0.37 \\
		Microwave, Kettle, Hinge Cabinet, Slide Cabinet & 2.45$\pm$0.25 & \textbf{3.15$\pm$0.22} & 1.85$\pm$0.23 \\
		Microwave, Kettle, Hinge Cabinet, Slide Cabinet & 2.01$\pm$0.24 & \textbf{2.95$\pm$0.44} & 2.44$\pm$0.47 \\
		\bottomrule
	\end{tabular}%
	\label{tab:ablationofkofkitchen}%
\end{table}%

\subsection{Effects of Fine-tuning Steps}

As all few-shot learning problems, the fine-tuning steps in new tasks to some extent determine the performance of the trained model. It controls the balance between under-fitting and over-fitting. We perform ablation studies of fine-tune steps in Meta-world benchmarks with $K=5$ and lr=1e-2 in  \ref{tab:ablationoffinetunestep}. Results are as follows:

\begin{table}[h!]
	\centering
	\caption{Ablation studies of the fine-tuning steps in Meat-world experiments with $ K=5 $.}
	\begin{tabular}{c|cccc|cccc}
		\toprule
		& \multicolumn{4}{c|}{ML10}     & \multicolumn{4}{c}{ML45} \\
		\cmidrule{2-9}          & \multicolumn{2}{c}{Meta-training} & \multicolumn{2}{c|}{Meta-testing} & \multicolumn{2}{c}{Meta-training} & \multicolumn{2}{c}{Meta-testing} \\
		fine-tune steps & 1-shot & 3-shot & 1-shot & 3-shot & 1-shot & 3-shot & 1-shot & 3-shot \\
		\midrule
		10    & 0.5   & 0.575 & \textbf{0.28} & 0.24  & 0.374 & 0.49  & 0.05  & 0.17 \\
		30    & 0.69  & \textbf{0.915} & 0.25  & 0.31  & 0.602 & 0.85  & 0.13  & 0.32 \\
		50    & \textbf{0.695} & 0.895 & 0.27  & 0.39  & 0.583 & 0.87  & 0.12  & 0.32 \\
		100   & 0.665 & 0.905 & 0.23  & 0.41  & 0.614 & 0.872 & 0.07  & 0.43 \\
		300   & 0.66  & 0.845 & \textbf{0.28} & \textbf{0.44} & \textbf{0.605} & \textbf{0.876} & 0.12  & \textbf{0.44} \\
		500   & 0.63  & 0.91  & 0.25  & 0.39  & 0.584 & 0.867 & \textbf{0.13} & 0.42 \\
		\midrule
		Range & 0.195 & 0.34  & 0.05  & 0.2   & 0.231 & 0.386 & 0.08  & 0.27 \\
		\bottomrule
	\end{tabular}%
	\label{tab:ablationoffinetunestep}%
\end{table}%

\subsection{Effects of Continuity Regularization}

We perform ablation studies of the effect of continuity regularization as following \ref{tab:ablationofconti}. DMIL\_nc means no continuity regularization. We can see that the continuity constraint would damage meta-training performance slightly, but increase the meta-testing performance greatly.

\begin{table}[h]
	\centering
	\caption{$ K=10 $}
	\begin{tabular}{c|cccc|cccc}
		\toprule
		& \multicolumn{4}{c|}{ML10}     & \multicolumn{4}{c}{ML45} \\
		\cmidrule{2-9}          & \multicolumn{2}{c}{Meta-training} & \multicolumn{2}{c|}{Meta-testing} & \multicolumn{2}{c}{Meta-training} & \multicolumn{2}{c}{Meta-testing} \\
		variants & 1-shot & 3-shot & 1-shot & 3-shot & 1-shot & 3-shot & 1-shot & 3-shot \\
		\midrule
		DMIL  & \textbf{0.795} & 0.94  & \textbf{0.52} & \textbf{0.57} & \textbf{0.713} & 0.92  & \textbf{0.21} & \textbf{0.48} \\
		DMIL\_nc & 0.788 & \textbf{0.96} & 0.32  & 0.56  & 0.703 & \textbf{0.927} & 0.17  & 0.35 \\
		\midrule
		Gap   & 0.007 & -0.02 & 0.2   & 0.01  & 0.01  & -0.007 & 0.04  & 0.13 \\
		\bottomrule
	\end{tabular}%
	\label{tab:ablationofconti}%
\end{table}%

\subsection{Effects of Hard/Soft EM Choices}

In DMIL, we use hard EM algorithm to train the high-level network. One may think about to use soft cross entropy loss to train the high-level network to get better results. We perform this ablation study in the following  \ref{tab:ablationofhardsoft}. We can see that a soft cross entropy training won't help increase the whole success rates. This may comes from that, usually we use soft cross entropy (such as label smoothing) to prevent over-fitting. However, in our situation, this may cause under-fitting, since training on such a large scale of diverse manipulation tasks is already very difficult. Future works can seek more comparisons about this choice.

\begin{table}[h]
	\centering
	\caption{Ablation about hard/soft EM choices with $ K=5 $ in the Meta-world environments.}
	\begin{tabular}{c|cccc|cccc}
		\toprule
		& \multicolumn{4}{c|}{ML10}     & \multicolumn{4}{c}{ML45} \\
		\cmidrule{2-9}          & \multicolumn{2}{c}{Meta-training} & \multicolumn{2}{c|}{Meta-testing} & \multicolumn{2}{c}{Meta-training} & \multicolumn{2}{c}{Meta-testing} \\
		variants & 1-shot & 3-shot & 1-shot & 3-shot & 1-shot & 3-shot & 1-shot & 3-shot \\
		\midrule
		DMIL  & \textbf{0.795} & \textbf{0.94} & \textbf{0.52} & \textbf{0.57} & \textbf{0.713333} & \textbf{0.92} & \textbf{0.21} & \textbf{0.48} \\
		DMIL\_soft & 0.37  & 0.65  & 0.33  & 0.46  & 0.235556 & 0.43  & 0.1   & 0.32 \\
		\midrule
		Gap   & 0.425 & 0.29  & 0.19  & 0.11  & 0.477778 & 0.49  & 0.11  & 0.16 \\
		\bottomrule
	\end{tabular}%
	\label{tab:ablationofhardsoft}%
\end{table}%

\begin{figure}[h]
	\centering
	\includegraphics[width=0.8\linewidth]{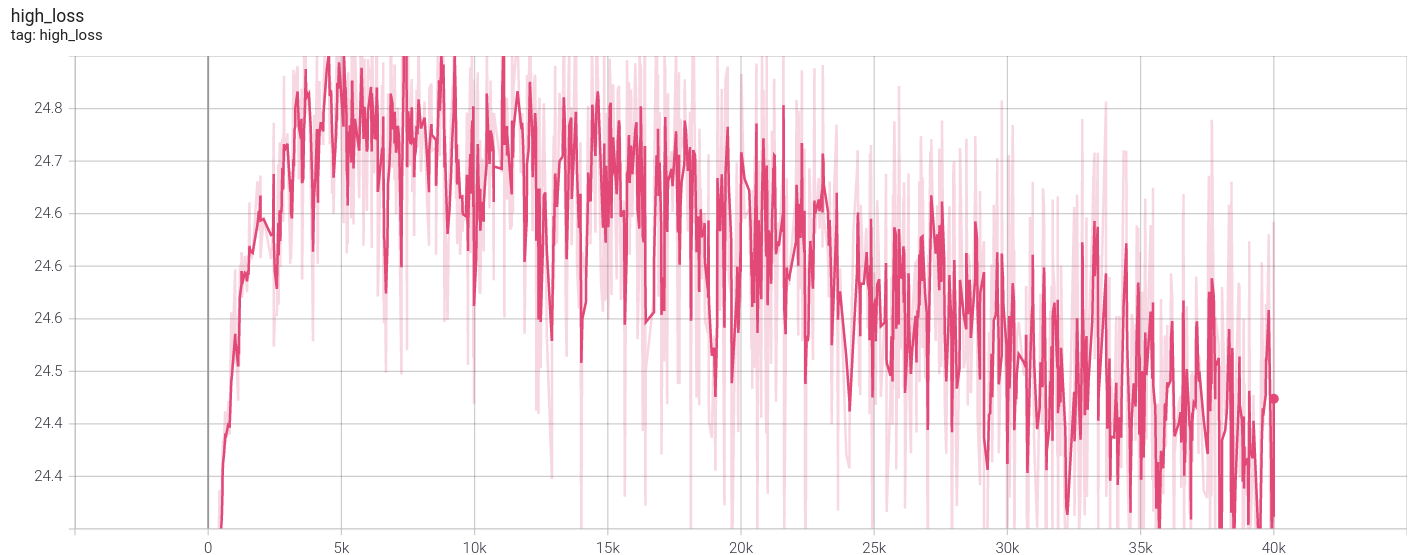}
	\caption{The training loss of the high-level network with a softmax shows a trend of rising first and then falling.\label{highloss}}
\end{figure}

Additionally, we found an interesting phenomena that the training loss of the high-level network with a softmax shows a trend of rising first and then falling, as shown in \ref{highloss}. In our experiments, a softmax loss may regularize the optimization process to make the high-level network be under-fitting on training data. This may come from that, the experiment environment (Meta-world) contains a large scale of manipulation tasks, in which the training of the high-level network can be difficult and unstable. Thus a soft-max cross entropy loss cannot help that much here  like how it works as a regularizer to prevent over-fitting in the label smoothing \cite{labelsmoothing}.

\section{Experiment Details}
\label{appendix:experiment}

\subsection{Environments}

\begin{figure}[h!]
	\centering
	\includegraphics[width=\linewidth]{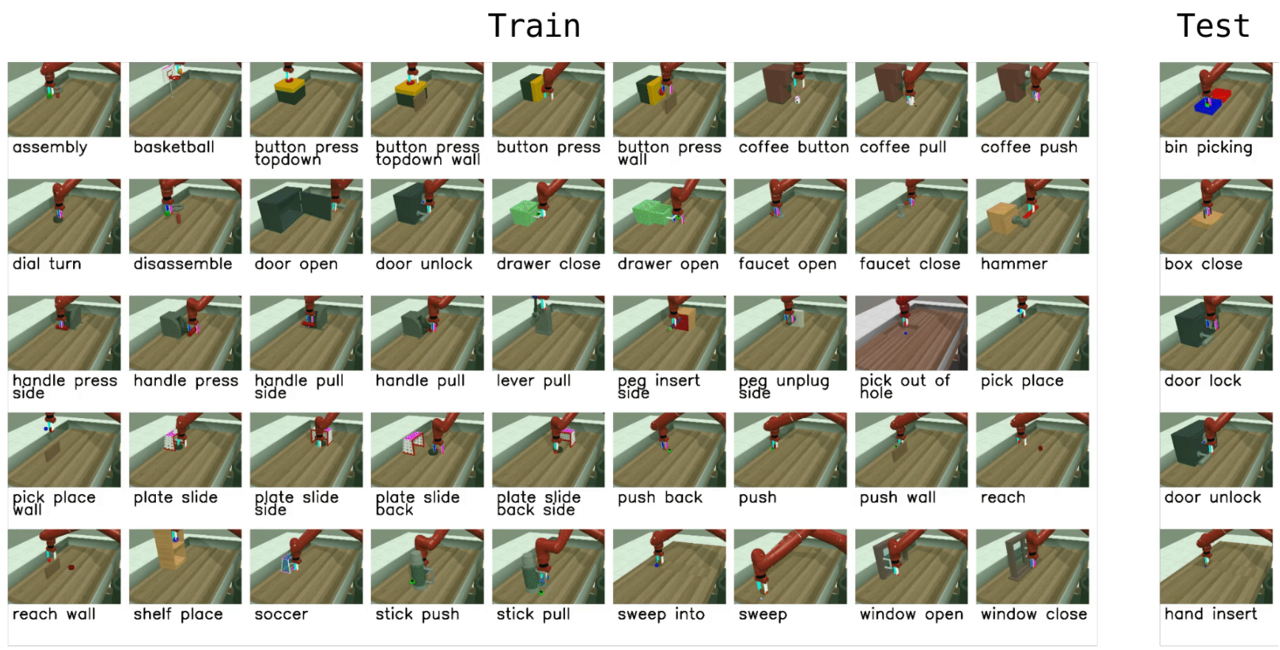}
	
	\caption{The ML45 environment.\label{ml45}}
\end{figure}

\begin{figure}[h!]
	\centering
	\includegraphics[width=\linewidth]{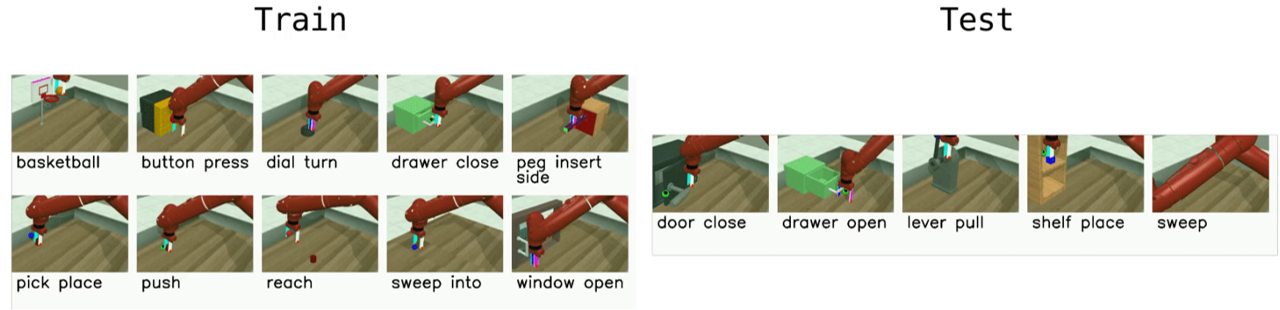}
	
	\caption{The ML10 environment.\label{ml10}}
\end{figure}

\begin{figure}[h]
	
	\centering
	\includegraphics[width=\linewidth]{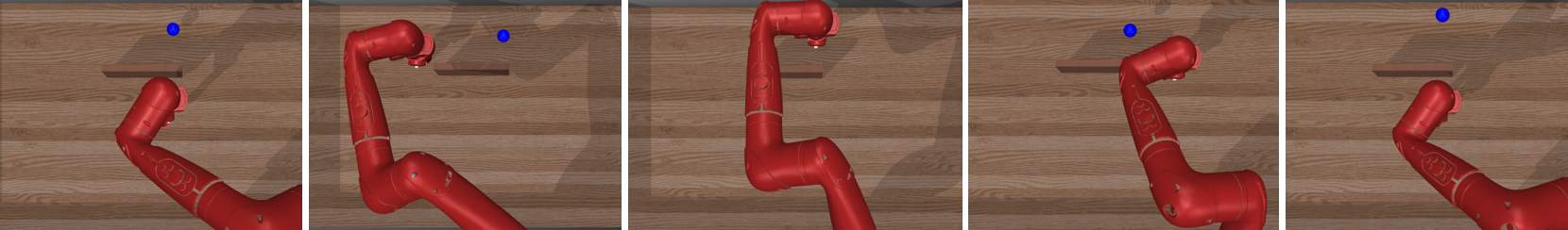}
	\caption{Task \textit{push-around-wall}.\label{pusharoundwall}}
\end{figure}

\begin{figure}[h]
	\centering
	\includegraphics[width=\linewidth]{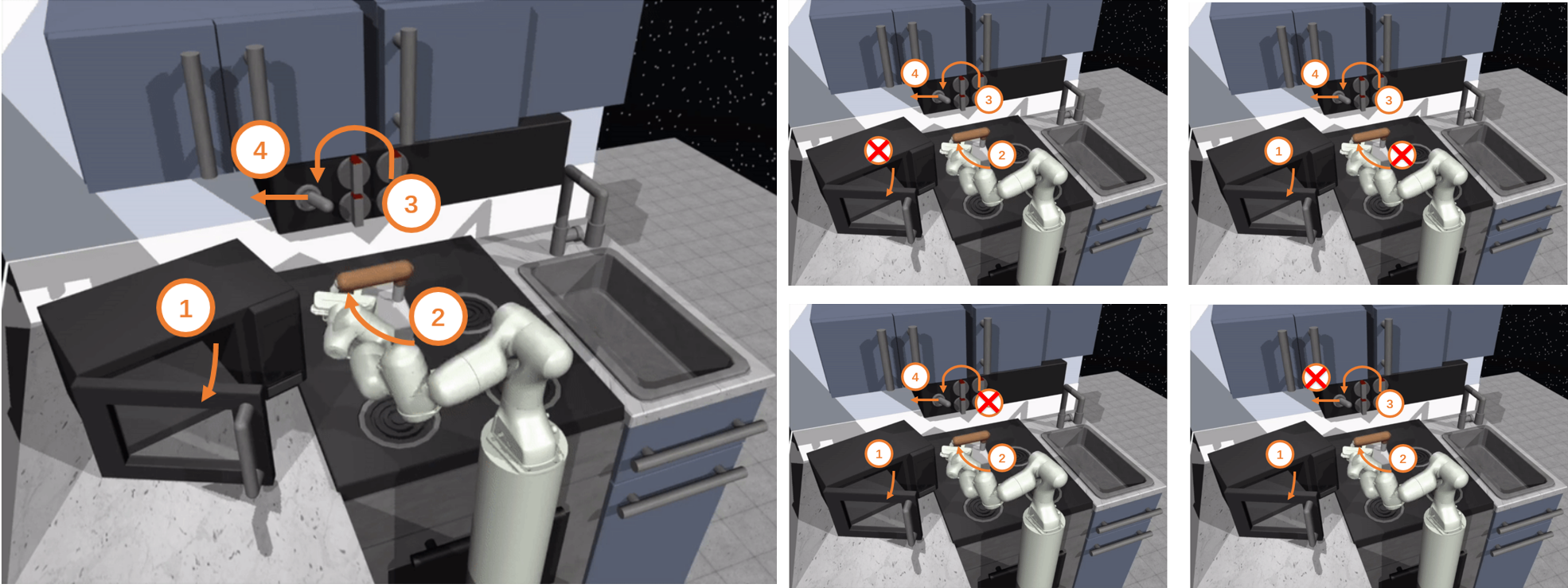}
	\caption{Kitchen environments.\label{kitchenenv}}
\end{figure}

See \ref{ml10},\ref{ml45},\ref{kitchenenv} and \ref{pusharoundwall}.

\subsection{Model Setup}
\label{modelsetup}

\textbf{DMIL:} The high-level network and each sub-skill is modeled with a 4-layer fully-connected neuron network, with 512 ReLU units in each layer. We use Adam as the meta-optimizer.  \textbf{DMIL-High} and \textbf{DMIL-Low} use the same architecture with DMIL. Hyper-parameter settings are available in table \ref{tab:DMIL}.

\begin{table}[!t]
	\centering
	\caption{DMIL hyper-parameters.}
	\begin{tabular}{|c|c|}
		\hline
		\textbf{Parameter} & \textbf{Value} \\
		\hline $ \alpha $
		& 5e-4 \\
		\hline $ \beta $
		& 1e-4 \\
		\hline 
		fine-tune iterations & 3 \\
		\hline
		batch size (in trajectory) & 16 \\
		\hline
		$ \lambda $     & 0.1 \\
		\hline
	\end{tabular}%
	\label{tab:DMIL}%
\end{table}%

\textbf{MIL:} We use a transformer \cite{transformer} as the policy to perform MAML. The input of the encoder is the whole one-shot demonstration or 3-shot demonstrations with concatenated state and action. The input of the decoder is current state. The output is the predicted action. Hyper-parameter settings are available in  \ref{tab:MIL}.

\begin{table}[h]
	\centering
	\caption{MIL hyper-parameters.}
	\begin{tabular}{|c|c|}
		\hline
		\textbf{Parameter} & \textbf{Value} \\
		\hline $ n_{head} $
		& 8 \\
		\hline $ n_{layer} $
		& 3 \\
		\hline $ d_{model} $
		& 512 \\
		\hline $ d_{k} $
		& 64 \\
		\hline $ d_{v} $
		& 64 \\
		\hline $ n_{position} $
		& 250 \\
		\hline dropout
		& 0.1 \\
		\hline
		batch size (in state-action pair) & 512 \\
		\hline
	\end{tabular}%
	\label{tab:MIL}%
\end{table}%

\textbf{MLSH:} We use the same settings of network with DMIL here. The macro step of high-level network is 3. Since our problem is not a reinforcement learning process, we use the \textit{behavior cloning} variant of MLSH. The pseudo reward is defined by the negative mean square loss of the predicted action and the ground truth, and we perform pseudo reinforcement learning process with off-policy demonstration data. We use PPO as our reinforcement learning algorithm. Note in this way we ignore the importance sampling weights that required by replacing the sampling process in the environments with the demonstrations in the replay buffer, which has been shown to be effective in practice in \cite{offpolicy,smile}. Hyper-parameter settings are available in  \ref{tab:MLSH}.

\begin{table}[h]
	\centering
	\caption{MLSH hyper-parameters.}
	\begin{tabular}{|c|c|}
		\hline
		\textbf{Parameter} & \textbf{Value} \\
		\hline high-level learning rate
		& 1e-3 \\
		\hline sub-skill learning rate
		& 1e-4 \\
		\hline PPO clip threshold
		& 0.02 \\
		\hline high-level warmup step
		& 500 \\
		\hline joint update step
		& 1000 \\
		\hline
		batch size (in state-action pair) & 900 \\
		\hline
	\end{tabular}%
	\label{tab:MLSH}%
\end{table}%

\textbf{PEMIRL:} We use the same setting of the high-level network as the policy $ \pi_\omega $ and the inference model $ q_\psi $ in PEMIRL. We use PPO as our reinforcement learning algorithm. We use a 3-layer fully-connected neuron network as the context-dependent disentangled reward estimator $ r_\theta(s,m) $ and the context-dependent potential function $ h_\phi(s,m) $. Here we also use the \textit{behavior cloning} variant of PEMIRL to only train models on the off-policy data. Hyper-parameter settings are available in   \ref{tab:PEMIRL}.

\begin{table}[htbp]
	\centering
	\caption{PEMIRL hyper-parameters.}
	\begin{tabular}{|c|c|}
		\hline
		\textbf{Parameter} & \textbf{Value} \\
		\hline learning rate of all models
		& 1e-4 \\
		\hline PPO clip threshold
		& 0.02 \\
		\hline coefficient ($ \gamma $) of $ h_\phi $ 
		& 1 \\
		\hline $ \beta $
		& 0.1 \\
		\hline
		batch size (in trajectory) & 16 \\
		\hline
	\end{tabular}%
	\label{tab:PEMIRL}%
\end{table}%

\subsection{Training Details}

For fine-tuning, OptionGAIL, DMIL-Low and DMIL-High have no meta-learning mechanism for (some parts of) the trained model. In the few-shot adaptation process, we have different fine-tune method for these baselines:

\textbf{OptionGAIL:} We train OptionGAIL models on the provided few-shot demonstrations for a few epochs.

\textbf{DMIL-Low:} We fix the high-level network and only fine-tune sub-skills with few-shot demonstrations.

\textbf{DMIL-High}: We fix sub-skills and only fine-tune the high-level network with few-shot demonstrations.

\end{document}

%% file: math_commands.tex
\usepackage{amsmath,amsfonts,bm}

\def\eqref#1{equation~\ref{#1}}

\def\1{\bm{1}}

\DeclareMathAlphabet{\mathsfit}{\encodingdefault}{\sfdefault}{m}{sl}
\SetMathAlphabet{\mathsfit}{bold}{\encodingdefault}{\sfdefault}{bx}{n}













%% file: iclr2022_conference.bbl
\begin{thebibliography}{56}
\providecommand{\natexlab}[1]{#1}
\providecommand{\url}[1]{\texttt{#1}}
\expandafter\ifx\csname urlstyle\endcsname\relax
  \providecommand{\doi}[1]{doi: #1}\else
  \providecommand{\doi}{doi: \begingroup \urlstyle{rm}\Url}\fi

\bibitem[Alet et~al.(2018)Alet, Lozano{-}P{\'{e}}rez, and
  Kaelbling]{modularmetalearning}
Ferran Alet, Tom{\'{a}}s Lozano{-}P{\'{e}}rez, and Leslie~Pack Kaelbling.
\newblock Modular meta-learning.
\newblock In \emph{2nd Annual Conference on Robot Learning}, 2018.

\bibitem[Atkeson \& Schaal(1997)Atkeson and Schaal]{bc}
Christopher~G. Atkeson and Stefan Schaal.
\newblock Robot learning from demonstration.
\newblock In \emph{Proceedings of the Fourteenth International Conference on
  Machine Learning}, 1997.

\bibitem[Cachet et~al.(2020)Cachet, Perez, and Kim]{transformerbasedmeta}
T~Cachet, J~Perez, and S~Kim.
\newblock Transformer-based meta-imitation learning for robotic manipulation.
\newblock In \emph{Neural Information Processing Systems, Workshop on Robot
  Learning}, 2020.

\bibitem[Chen et~al.(2016)Chen, Duan, Houthooft, Schulman, Sutskever, and
  Abbeel]{infogan}
Xi~Chen, Yan Duan, Rein Houthooft, John Schulman, Ilya Sutskever, and Pieter
  Abbeel.
\newblock Infogan: Interpretable representation learning by information
  maximizing generative adversarial nets.
\newblock In \emph{Advances in Neural Information Processing Systems}, 2016.

\bibitem[Daniel et~al.(2016)Daniel, van Hoof, Peters, and Neumann]{ddo}
Christian Daniel, Herke van Hoof, Jan Peters, and Gerhard Neumann.
\newblock Probabilistic inference for determining options in reinforcement
  learning.
\newblock \emph{Machine Learning}, 2016.

\bibitem[Duan et~al.(2017)Duan, Andrychowicz, Stadie, Ho, Schneider, Sutskever,
  Abbeel, and Zaremba]{oneshotimitationlearning}
Yan Duan, Marcin Andrychowicz, Bradly~C. Stadie, Jonathan Ho, Jonas Schneider,
  Ilya Sutskever, Pieter Abbeel, and Wojciech Zaremba.
\newblock One-shot imitation learning.
\newblock In \emph{Advances in Neural Information Processing Systems}, 2017.

\bibitem[Duvenaud et~al.(2016)Duvenaud, Maclaurin, and Adams]{earlystopping}
David Duvenaud, Dougal Maclaurin, and Ryan Adams.
\newblock Early stopping as nonparametric variational inference.
\newblock In \emph{Artificial Intelligence and Statistics}. PMLR, 2016.

\bibitem[Fallah et~al.(2020)Fallah, Mokhtari, and Ozdaglar]{convergenceofmaml}
Alireza Fallah, Aryan Mokhtari, and Asuman~E. Ozdaglar.
\newblock On the convergence theory of gradient-based model-agnostic
  meta-learning algorithms.
\newblock In \emph{The 23rd International Conference on Artificial Intelligence
  and Statistics}, 2020.

\bibitem[Finn et~al.(2017{\natexlab{a}})Finn, Abbeel, and Levine]{maml}
Chelsea Finn, Pieter Abbeel, and Sergey Levine.
\newblock Model-agnostic meta-learning for fast adaptation of deep networks.
\newblock In \emph{Proceedings of the 34th International Conference on Machine
  Learning}, 2017{\natexlab{a}}.

\bibitem[Finn et~al.(2017{\natexlab{b}})Finn, Yu, Zhang, Abbeel, and
  Levine]{oneshotimitationvisuallearningviametalearning}
Chelsea Finn, Tianhe Yu, Tianhao Zhang, Pieter Abbeel, and Sergey Levine.
\newblock One-shot visual imitation learning via meta-learning.
\newblock In \emph{1st Annual Conference on Robot Learning},
  2017{\natexlab{b}}.

\bibitem[Finn et~al.(2018)Finn, Xu, and Levine]{pmaml}
Chelsea Finn, Kelvin Xu, and Sergey Levine.
\newblock Probabilistic model-agnostic meta-learning.
\newblock In \emph{Advances in Neural Information Processing Systems}, 2018.

\bibitem[Florensa et~al.(2017)Florensa, Duan, and Abbeel]{stochasticnn}
Carlos Florensa, Yan Duan, and Pieter Abbeel.
\newblock Stochastic neural networks for hierarchical reinforcement learning.
\newblock In \emph{5th International Conference on Learning Representations},
  2017.

\bibitem[Frans et~al.(2018)Frans, Ho, Chen, Abbeel, and
  Schulman]{metalearningsharedhierarchies}
Kevin Frans, Jonathan Ho, Xi~Chen, Pieter Abbeel, and John Schulman.
\newblock Meta learning shared hierarchies.
\newblock In \emph{6th International Conference on Learning Representations},
  2018.

\bibitem[Fu et~al.(2020)Fu, Kumar, Nachum, Tucker, and Levine]{d4rl}
Justin Fu, Aviral Kumar, Ofir Nachum, George Tucker, and Sergey Levine.
\newblock D4rl: Datasets for deep data-driven reinforcement learning.
\newblock \emph{arXiv preprint arXiv:2004.07219}, 2020.

\bibitem[Ghasemipour et~al.(2019)Ghasemipour, Gu, and Zemel]{smile}
Seyed Kamyar~Seyed Ghasemipour, Shixiang Gu, and Richard~S. Zemel.
\newblock Smile: Scalable meta inverse reinforcement learning through
  context-conditional policies.
\newblock In \emph{Advances in Neural Information Processing Systems 32:},
  2019.

\bibitem[Grant et~al.(2018)Grant, Finn, Levine, Darrell, and
  Griffiths]{recasting}
Erin Grant, Chelsea Finn, Sergey Levine, Trevor Darrell, and Thomas~L.
  Griffiths.
\newblock Recasting gradient-based meta-learning as hierarchical bayes.
\newblock In \emph{6th International Conference on Learning Representations},
  2018.

\bibitem[Gupta et~al.(2019)Gupta, Kumar, Lynch, Levine, and
  Hausman]{replaypolicylearning}
Abhishek Gupta, Vikash Kumar, Corey Lynch, Sergey Levine, and Karol Hausman.
\newblock Relay policy learning: Solving long-horizon tasks via imitation and
  reinforcement learning.
\newblock \emph{arXiv preprint arXiv:1910.11956}, 2019.

\bibitem[Haarnoja et~al.(2018)Haarnoja, Hartikainen, Abbeel, and
  Levine]{latentspace}
Tuomas Haarnoja, Kristian Hartikainen, Pieter Abbeel, and Sergey Levine.
\newblock Latent space policies for hierarchical reinforcement learning.
\newblock In Jennifer~G. Dy and Andreas Krause (eds.), \emph{Proceedings of the
  35th International Conference on Machine Learning}, 2018.

\bibitem[Hakhamaneshi et~al.(2021)Hakhamaneshi, Zhao, Zhan, Abbeel, and
  Laskin]{fist}
Kourosh Hakhamaneshi, Ruihan Zhao, Albert Zhan, Pieter Abbeel, and Michael
  Laskin.
\newblock Hierarchical few-shot imitation with skill transition models.
\newblock 2021.

\bibitem[Hausknecht \& Stone(2016)Hausknecht and Stone]{moe1}
Matthew~J. Hausknecht and Peter Stone.
\newblock Deep reinforcement learning in parameterized action space.
\newblock In \emph{4th International Conference on Learning Representations},
  2016.

\bibitem[Hausman et~al.(2018)Hausman, Springenberg, Wang, Heess, and
  Riedmiller]{learninganembeddingspace}
Karol Hausman, Jost~Tobias Springenberg, Ziyu Wang, Nicolas Heess, and
  Martin~A. Riedmiller.
\newblock Learning an embedding space for transferable robot skills.
\newblock In \emph{6th International Conference on Learning Representations},
  2018.

\bibitem[Henderson et~al.(2018)Henderson, Chang, Bacon, Meger, Pineau, and
  Precup]{optiongan}
Peter Henderson, Wei{-}Di Chang, Pierre{-}Luc Bacon, David Meger, Joelle
  Pineau, and Doina Precup.
\newblock Optiongan: Learning joint reward-policy options using generative
  adversarial inverse reinforcement learning.
\newblock In \emph{Proceedings of the Thirty-Second {AAAI} Conference on
  Artificial Intelligence}, 2018.

\bibitem[Ho \& Ermon(2016)Ho and Ermon]{gail}
Jonathan Ho and Stefano Ermon.
\newblock Generative adversarial imitation learning.
\newblock In \emph{Advances in Neural Information Processing Systems}, 2016.

\bibitem[Jacobs et~al.(1991)Jacobs, Jordan, Nowlan, and Hinton]{moe3}
Robert~A. Jacobs, Michael~I. Jordan, Steven~J. Nowlan, and Geoffrey~E. Hinton.
\newblock Adaptive mixtures of local experts.
\newblock \emph{Neural Computing}, 1991.

\bibitem[Jing et~al.(2021)Jing, Huang, Sun, Ma, Kong, Gan, and Li]{optiongail}
Mingxuan Jing, Wenbing Huang, Fuchun Sun, Xiaojian Ma, Tao Kong, Chuang Gan,
  and Lei Li.
\newblock Adversarial option-aware hierarchical imitation learning.
\newblock In \emph{Proceedings of the 38th International Conference on Machine
  Learning}, 2021.

\bibitem[Kostrikov et~al.(2018)Kostrikov, Agrawal, Levine, and
  Tompson]{offpolicy}
Ilya Kostrikov, Kumar~Krishna Agrawal, Sergey Levine, and Jonathan Tompson.
\newblock Addressing sample inefficiency and reward bias in inverse
  reinforcement learning.
\newblock \emph{arXiv preprint arXiv:1809.02925}, 2018.

\bibitem[Krishnan et~al.(2017)Krishnan, Fox, Stoica, and Goldberg]{ddco}
Sanjay Krishnan, Roy Fox, Ion Stoica, and Ken Goldberg.
\newblock {DDCO:} discovery of deep continuous options for robot learning from
  demonstrations.
\newblock In \emph{1st Annual Conference on Robot Learning}, 2017.

\bibitem[Le et~al.(2018)Le, Jiang, Agarwal, Dud{\'{\i}}k, Yue, and
  III]{hierarchicalimitationlearning}
Hoang~Minh Le, Nan Jiang, Alekh Agarwal, Miroslav Dud{\'{\i}}k, Yisong Yue, and
  Hal~Daum{\'{e}} III.
\newblock Hierarchical imitation and reinforcement learning.
\newblock In \emph{Proceedings of the 35th International Conference on Machine
  Learning}, 2018.

\bibitem[Lee \& Seo(2020)Lee and
  Seo]{learningcompoundtaskswithouttaskspecificknowledge}
Sang{-}Hyun Lee and Seung{-}Woo Seo.
\newblock Learning compound tasks without task-specific knowledge via imitation
  and self-supervised learning.
\newblock In \emph{Proceedings of the 37th International Conference on Machine
  Learning}, 2020.

\bibitem[Li et~al.(2017)Li, Song, and Ermon]{infogail}
Yunzhu Li, Jiaming Song, and Stefano Ermon.
\newblock Infogail: Interpretable imitation learning from visual
  demonstrations.
\newblock In \emph{Advances in Neural Information Processing Systems}, 2017.

\bibitem[Liu \& Hodgins(2017)Liu and Hodgins]{pretrain1}
Libin Liu and Jessica~K. Hodgins.
\newblock Learning to schedule control fragments for physics-based characters
  using deep q-learning.
\newblock \emph{{ACM} Trans. Graph.}, 2017.

\bibitem[Lynch et~al.(2019)Lynch, Khansari, Xiao, Kumar, Tompson, Levine, and
  Sermanet]{latentplay}
Corey Lynch, Mohi Khansari, Ted Xiao, Vikash Kumar, Jonathan Tompson, Sergey
  Levine, and Pierre Sermanet.
\newblock Learning latent plans from play.
\newblock In \emph{3rd Annual Conference on Robot Learning}, 2019.

\bibitem[Merel et~al.(2019)Merel, Ahuja, Pham, Tunyasuvunakool, Liu, Tirumala,
  Heess, and Wayne]{pretrain2}
Josh Merel, Arun Ahuja, Vu~Pham, Saran Tunyasuvunakool, Siqi Liu, Dhruva
  Tirumala, Nicolas Heess, and Greg Wayne.
\newblock Hierarchical visuomotor control of humanoids.
\newblock In \emph{7th International Conference on Learning Representations},
  2019.

\bibitem[M{\"u}ller et~al.(2019)M{\"u}ller, Kornblith, and
  Hinton]{labelsmoothing}
Rafael M{\"u}ller, Simon Kornblith, and Geoffrey Hinton.
\newblock When does label smoothing help?
\newblock \emph{arXiv preprint arXiv:1906.02629}, 2019.

\bibitem[Neumann et~al.(2009)Neumann, Maass, and Peters]{moe2}
Gerhard Neumann, Wolfgang Maass, and Jan Peters.
\newblock Learning complex motions by sequencing simpler motion templates.
\newblock In Andrea~Pohoreckyj Danyluk, L{\'{e}}on Bottou, and Michael~L.
  Littman (eds.), \emph{Proceedings of the 26th Annual International Conference
  on Machine Learning}, 2009.

\bibitem[Osa et~al.(2018)Osa, Pajarinen, Neumann, Bagnell, Abbeel, and
  Peters]{analtorithmicperspectiveonimitationlearning}
Takayuki Osa, Joni Pajarinen, Gerhard Neumann, J.~Andrew Bagnell, Pieter
  Abbeel, and Jan Peters.
\newblock An algorithmic perspective on imitation learning.
\newblock 2018.

\bibitem[Paszke et~al.(2019)Paszke, Gross, Massa, Lerer, Bradbury, Chanan,
  Killeen, Lin, Gimelshein, Antiga, et~al.]{pytorch}
Adam Paszke, Sam Gross, Francisco Massa, Adam Lerer, James Bradbury, Gregory
  Chanan, Trevor Killeen, Zeming Lin, Natalia Gimelshein, Luca Antiga, et~al.
\newblock Pytorch: An imperative style, high-performance deep learning library.
\newblock \emph{Advances in neural information processing systems}, 2019.

\bibitem[Peng et~al.(2018)Peng, Abbeel, Levine, and van~de Panne]{deepmimic}
Xue~Bin Peng, Pieter Abbeel, Sergey Levine, and Michiel van~de Panne.
\newblock Deepmimic: example-guided deep reinforcement learning of
  physics-based character skills.
\newblock \emph{{ACM} Trans. Graph.}, 2018.

\bibitem[Peng et~al.(2019)Peng, Chang, Zhang, Abbeel, and Levine]{mcp}
Xue~Bin Peng, Michael Chang, Grace Zhang, Pieter Abbeel, and Sergey Levine.
\newblock {MCP:} learning composable hierarchical control with multiplicative
  compositional policies.
\newblock In \emph{Advances in Neural Information Processing Systems}, 2019.

\bibitem[Price \& Boutilier(2003)Price and Boutilier]{spirl}
Bob Price and Craig Boutilier.
\newblock Accelerating reinforcement learning through implicit imitation.
\newblock \emph{Journal of Artificial Intelligence Research}, 19:\penalty0
  569--629, 2003.

\bibitem[Ravi \& Beatson(2019)Ravi and Beatson]{amortizedbayesianmeta}
Sachin Ravi and Alex Beatson.
\newblock Amortized bayesian meta-learning.
\newblock In \emph{7th International Conference on Learning Representations},
  2019.

\bibitem[Ross et~al.(2011)Ross, Gordon, and Bagnell]{areductionofimitation}
St{\'{e}}phane Ross, Geoffrey~J. Gordon, and Drew Bagnell.
\newblock A reduction of imitation learning and structured prediction to
  no-regret online learning.
\newblock In \emph{Proceedings of the Fourteenth International Conference on
  Artificial Intelligence and Statistics}, 2011.

\bibitem[Santos(1996)]{earlylinear}
Reginaldo~J Santos.
\newblock Equivalence of regularization and truncated iteration for general
  ill-posed problems.
\newblock \emph{Linear algebra and its applications}, 1996.

\bibitem[Sharma et~al.(2019)Sharma, Sharma, Rhinehart, and
  Kitani]{directedinfogail}
Mohit Sharma, Arjun Sharma, Nicholas Rhinehart, and Kris~M. Kitani.
\newblock Directed-info {GAIL:} learning hierarchical policies from unsegmented
  demonstrations using directed information.
\newblock In \emph{7th International Conference on Learning Representations},
  2019.

\bibitem[Shu et~al.(2018)Shu, Xiong, and Socher]{hierarchicalmultitask}
Tianmin Shu, Caiming Xiong, and Richard Socher.
\newblock Hierarchical and interpretable skill acquisition in multi-task
  reinforcement learning.
\newblock In \emph{6th International Conference on Learning Representations},
  2018.

\bibitem[Sutton et~al.(1999)Sutton, Precup, and Singh]{option}
Richard~S. Sutton, Doina Precup, and Satinder~P. Singh.
\newblock Between mdps and semi-mdps: {A} framework for temporal abstraction in
  reinforcement learning.
\newblock \emph{Artif. Intell.}, 1999.

\bibitem[Vaswani et~al.(2017)Vaswani, Shazeer, Parmar, Uszkoreit, Jones, Gomez,
  Kaiser, and Polosukhin]{transformer}
Ashish Vaswani, Noam Shazeer, Niki Parmar, Jakob Uszkoreit, Llion Jones,
  Aidan~N. Gomez, Lukasz Kaiser, and Illia Polosukhin.
\newblock Attention is all you need.
\newblock In \emph{Advances in Neural Information Processing Systems}, 2017.

\bibitem[Vuorio et~al.(2019)Vuorio, Sun, Hu, and Lim]{multimodalmaml}
Risto Vuorio, Shao{-}Hua Sun, Hexiang Hu, and Joseph~J. Lim.
\newblock Multimodal model-agnostic meta-learning via task-aware modulation.
\newblock In \emph{Advances in Neural Information Processing Systems}, 2019.

\bibitem[Xu et~al.(2019)Xu, Ratner, Dragan, Levine, and
  Finn]{learningaprioroverintent}
Kelvin Xu, Ellis Ratner, Anca~D. Dragan, Sergey Levine, and Chelsea Finn.
\newblock Learning a prior over intent via meta-inverse reinforcement learning.
\newblock In \emph{Proceedings of the 36th International Conference on Machine
  Learning}, 2019.

\bibitem[Yao et~al.(2019)Yao, Wei, Huang, and
  Li]{hierarchicallystructuredmetalearning}
Huaxiu Yao, Ying Wei, Junzhou Huang, and Zhenhui Li.
\newblock Hierarchically structured meta-learning.
\newblock In \emph{Proceedings of the 36th International Conference on Machine
  Learning}, 2019.

\bibitem[Yu et~al.(2019{\natexlab{a}})Yu, Yu, Finn, and Ermon]{metairl}
Lantao Yu, Tianhe Yu, Chelsea Finn, and Stefano Ermon.
\newblock Meta-inverse reinforcement learning with probabilistic context
  variables.
\newblock In \emph{Advances in Neural Information Processing Systems},
  2019{\natexlab{a}}.

\bibitem[Yu et~al.(2018{\natexlab{a}})Yu, Abbeel, Levine, and
  Finn]{oneshothierarchicalimitationlearning}
Tianhe Yu, Pieter Abbeel, Sergey Levine, and Chelsea Finn.
\newblock One-shot hierarchical imitation learning of compound visuomotor
  tasks.
\newblock \emph{arXiv preprint arXiv:1810.11043}, 2018{\natexlab{a}}.

\bibitem[Yu et~al.(2018{\natexlab{b}})Yu, Finn, Dasari, Xie, Zhang, Abbeel, and
  Levine]{oneshotimitationfromobservinghumans}
Tianhe Yu, Chelsea Finn, Sudeep Dasari, Annie Xie, Tianhao Zhang, Pieter
  Abbeel, and Sergey Levine.
\newblock One-shot imitation from observing humans via domain-adaptive
  meta-learning.
\newblock In \emph{Robotics: Science and Systems XIV}, 2018{\natexlab{b}}.

\bibitem[Yu et~al.(2019{\natexlab{b}})Yu, Quillen, He, Julian, Hausman, Finn,
  and Levine]{metaworld}
Tianhe Yu, Deirdre Quillen, Zhanpeng He, Ryan Julian, Karol Hausman, Chelsea
  Finn, and Sergey Levine.
\newblock Meta-world: {A} benchmark and evaluation for multi-task and meta
  reinforcement learning.
\newblock In \emph{3rd Annual Conference on Robot Learning},
  2019{\natexlab{b}}.

\bibitem[Zhu et~al.(2018)Zhu, Wang, Merel, Rusu, Erez, Cabi, Tunyasuvunakool,
  Kram{\'{a}}r, Hadsell, de~Freitas, and Heess]{reinforceandimitation}
Yuke Zhu, Ziyu Wang, Josh Merel, Andrei~A. Rusu, Tom Erez, Serkan Cabi, Saran
  Tunyasuvunakool, J{\'{a}}nos Kram{\'{a}}r, Raia Hadsell, Nando de~Freitas,
  and Nicolas Heess.
\newblock Reinforcement and imitation learning for diverse visuomotor skills.
\newblock In \emph{Robotics: Science and Systems XIV}, 2018.

\bibitem[Zou \& Lu(2020)Zou and Lu]{gradientem}
Yayi Zou and Xiaoqi Lu.
\newblock Gradient-em bayesian meta-learning.
\newblock In \emph{Advances in Neural Information Processing Systems 33: Annual
  Conference on Neural Information Processing Systems}, 2020.

\end{thebibliography}
